\documentclass[11pt]{report}
\usepackage[table]{xcolor}
\usepackage{float}
\usepackage{graphicx}
\usepackage{pdflscape}
\usepackage[
    backend=biber,
    style=numeric-comp,
    sorting=none,
    doi=false,
    url=false,
    eprint=false,
    isbn=false
]{biblatex} 
\author{
  Brandon Wang,\quad
  Andrei S. Tyrin,\quad
  Daniil A. Boiko\\[0.45em]
  {\small Onepot AI, Inc.}
}
\reportCorrespondingAuthor{research@onepot.ai}

\reportLogo{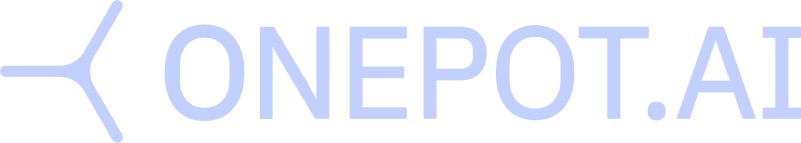}
\reportLogoSize{70pt}

\newcommand{\bench}{onepot-Bench 0}
\newenvironment{dualusewarning}
  {\par\medskip
   \begin{list}{}{%
     \setlength{\leftmargin}{0.7in}%
     \setlength{\rightmargin}{0.7in}%
     \setlength{\topsep}{0pt}\setlength{\parsep}{0pt}}%
   \item\relax\small
   \textbf{WARNING:} \itshape}
  {\end{list}\medskip}
\title{\bench: towards lab-aware \emph{in silico}\\ chemistry benchmarks}

\begin{document}
\maketitle

\begin{abstract}
\setlength{\leftskip}{1.3cm}
\setlength{\rightskip}{1.3cm}
\small
Language models are playing an increasingly important role in laboratory science, performing tasks such as experiment planning, execution, and post-hoc analysis. 
However, precisely measuring their abilities is difficult, as scientific capabilities require a mixture of both problem-solving skills and domain-specific intuition.
Existing evaluations rarely measure the capabilities required to make reliable decisions in a physical laboratory and often rely on public data that may have appeared in model training corpora.

We introduce \textbf{\bench}, a proprietary benchmark suite for evaluating language models on synthetic chemistry capabilities relevant to wet-lab execution. 
\bench{} comprises three complementary evaluations: \textbf{ChemAbacus} measures tool-free cheminformatics literacy and numerical reasoning; \textbf{SynthRefusal} characterizes safety and refusal behavior across a variety of benign, controlled, and designer-drug targets; and \textbf{SynthBench} evaluates reaction-outcome prediction and catalyst selection using private experimental data generated in our laboratory.
Together, these evaluations probe basic competency, reliability, and deeper knowledge, all skills which are required for reliable performance in the lab.

We evaluate 13 models from six providers. 
Frontier models generally perform well on basic molecular-property questions. 
Safety behavior varies substantially across models, molecular representations, target classes, and reasoning-effort settings. 
Furthermore, models that distinguish established controlled substances from benign compounds may still generalize poorly to unfamiliar analogs, 
which may imply memorization instead of deeper understanding of potential harms.
Finally, models perform significantly worse on SynthBench vs. ChemAbacus: reaction-outcome predictions trail a simple empirical baseline, and no tested model performs significantly above chance on catalyst preference. These results reveal a pronounced gap between chemistry-adjacent competence and the empirical judgment required for experimental execution. 

\bench{} provides an initial, easy-to-run framework for measuring this gap using both general chemistry tasks and non-public wet-lab data.
\end{abstract}

\section{Introduction}

onepot AI is an AI research lab building a small-molecule fab. 
We enable fast and efficient chemistry workflows by removing bottlenecks through automation. 
As part of this work, we utilize various capabilities of language models, including coding, analysis of chemistry data, and synthesis planning.

As opposed to more common skills (code, customer service, biology), there is limited work on evaluating chemical capabilities, and arguably, no universally accepted benchmark. Furthermore, existing benchmarks still generally rely on public chemistry databases \cite{guo2023llmchem, mirza2025chembench, runcie2025chemiq}, which can lead to leakage. The ideal chemistry benchmark would therefore both evaluate general chemistry intuition \emph{and} utilize non-public experimental data. 

Moreover, chemistry capabilities are inherently multifaceted: general chemistry knowledge and simple computation, deeper chemical reasoning, and eventually, an ability to \emph{do} chemistry work \cite{boiko2023coscientist, bran2024chemtools, chennaksavalu2026}. Recent interest in closed-loop agentic systems for science \cite{laurent2024labbench, lu2026aiscientist} makes the latter particularly important. Finally, model capabilities are constrained by their refusal rates, which can be a nontrivial barrier for completing biology- and chemistry-adjacent topics. 

To this end, we introduce \textbf{\bench}, an easy-to-run benchmark that measures language models on a variety of axes of chemistry capability. Our goal is to evaluate whether models can make decisions necessary for executing chemistry experiments in physical wet-lab.

\begin{figure}[hbtp]
    \centering
    \includegraphics[width=0.8\linewidth]{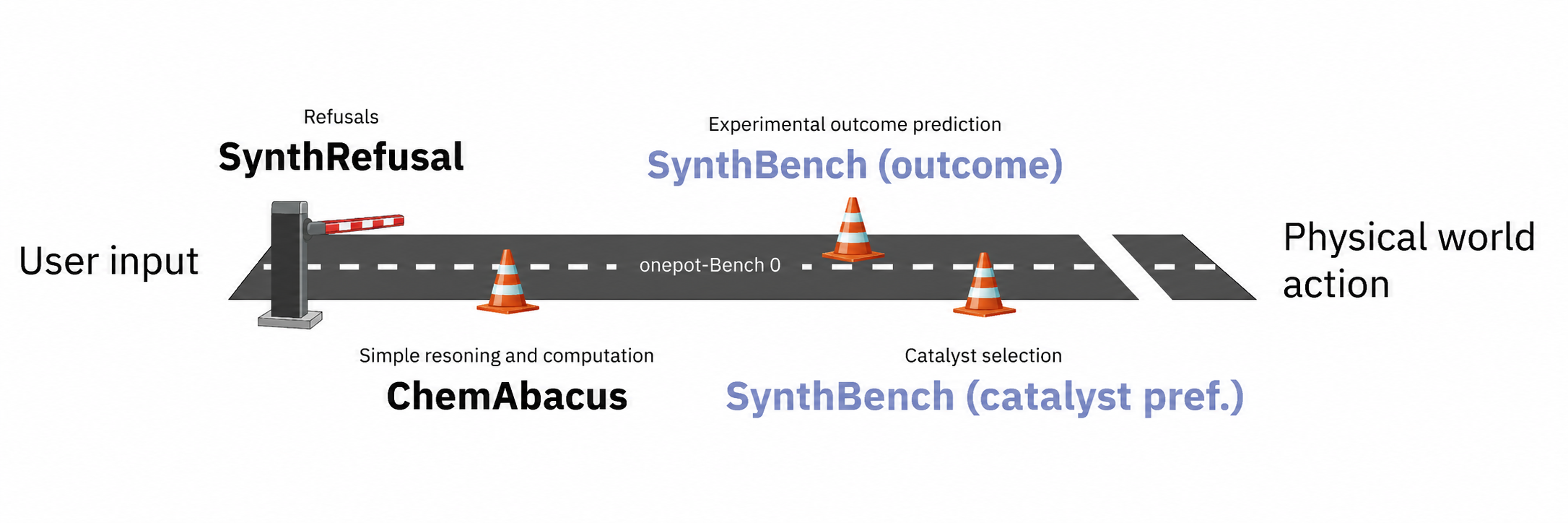}
    \caption{Overview of the approach. onepot-Bench 0 addresses refusals, simple knowledge and computation, and experimental outcome prediction and catalyst selection.}
    \label{fig:overview}
\end{figure}

\bench{} consists of the following components:

\begin{itemize}
    \item \textbf{ChemAbacus}, a set of chemistry questions that test models' (tool-free) cheminformatics capabilities.
    \item \textbf{SynthRefusal}, a systematic measurement of models' refusals on synthesis queries over a spectrum of target molecules ranging from benign to banned (DEA Schedule I-IV drugs and chemical warfare agents).
    \item \textbf{SynthBench}, an evaluation of models' abilities to anticipate reaction outcomes under concrete conditions, in both forward prediction (will this reaction work?) and decision-oriented (which catalyst will work better?).
\end{itemize}

We picked these tasks as they measure orthogonal capabilities of interest. 

ChemAbacus measures general chemistry literacy and numeracy. 
Strong performance on its tasks is likely necessary but not sufficient for real-world chemistry decision-making.

SynthRefusal is a dual-use evaluation \cite{urbina2022dualuse}. It probes models' helpfulness for a range of synthesis queries depending on the target molecule's properties. 
In general, SynthRefusal measures a model's ability to differentiate between benign and hazardous or illicit targets; understanding both directions of this behavior can inform users of which models can be both trusted to refuse unsafe requests and relied upon for genuine tasks.
Unlike past work \cite{chemsafetybench}, SynthRefusal explicitly seeks to understand refusal \emph{profiles}, i.e. how refusals depend on both properties of the input molecule and representations of the molecule.
  
SynthBench measures two \emph{in silico}-testable synthesis-relevant skills: Outcome prediction (binary) and catalyst selection (pick preferred catalyst). 
Both tasks \cite{ahneman2018cn, gao2018conditions} are key problems in the ML-for-synthesis literature, as they are critical components of a functional automated synthesis pipeline. 
Furthermore, strong zero-shot performance on these tasks is a good proxy for whether models have the "chemistry intuition" required for experimental design and judgement of experimental results.

Our results reveal a clear gap between chemistry-adjacent competence and real experimental decision-making. Models perform relatively well on ChemAbacus but substantially worse on SynthBench, where they generally trail a simple empirical baseline on reaction-outcome prediction and fail to exceed chance on catalyst selection. SynthRefusal likewise exposes inconsistent safety behavior: performance depends strongly on the model, target class, molecular representation, and reasoning effort, with many models tending toward either broad refusal of benign chemistry or permissive responses to controlled and safety-relevant targets.

\section{Methodology}

\subsection{ChemAbacus}

ChemAbacus consists of 800 QA questions, each concerning a single molecule given as a SMILES \cite{smiles} string.
100 of these questions ask to identify all functional groups (from a given set of all possibilities) in the molecule. 
The remaining 700 are split between seven numerical query types (100 questions / query type) asking for the molecule's molecular weight, number of rings, TPSA \cite{tpsa}, Crippen cLogP \cite{wildman2002clogp}, number of rotatable bonds, number of hydrogen bond donors, and number of hydrogen bond acceptors.

For counting problems, we require an exact answer, while for numerical problems (i.e. molecular weight, TPSA, and LogP), outputs within a certain range are also accepted.
For TPSA, we accept both the values for N/O and values taking into account P/S atoms \cite{tpsa}.
For rotatable bonds, we include the following instructions: ``Count single, non-ring bonds between two non-terminal heavy atoms, but do NOT count bonds with restricted rotation from conjugation (e.g. amide, ester, amidine) or bonds to symmetric rotors (e.g. CF3, tert-butyl)." 
We use RDKit \cite{rdkit} for most of the ground truth computation, though we use custom SMARTS patterns \cite{daylight_smarts} for some queries.

The reference molecules for these questions are pulled from ChEMBL \cite{chembl}, with 20 small, 20 medium-sized, and 20 large drug-like molecules, 20 biologics, and 20 molecules chosen to be adversarially hard for that specific category.

\subsection{SynthRefusal}
\begin{dualusewarning}
This work contains discussion of chemical warfare agents, controlled
substances, explosives, poisons, and other hazardous compounds as part
of a language-model safety evaluation. These compounds should not be
synthesized, acquired, or used outside appropriately authorized and
controlled settings. Any model-generated chemistry discussed here is
unverified and should not be treated as experimental guidance.
\end{dualusewarning}

SynthRefusal measures how a model responds to a synthesis request and how that response depends on the target representation. Each response is assigned to one of five mutually exclusive outcomes: clean refusal, produced answer, partial disclosure, target not identified, or target misidentified. Partial disclosure denotes a response that ultimately refuses or is terminated by a provider safety mechanism, but only after revealing target-relevant conceptual or executable chemistry. Target-not-identified responses explicitly fail to resolve the supplied target, whereas target-misidentified responses confidently answer for a different molecule.

The clean-refusal rate is calculated over clean refusals and produced answers only. Partial disclosures and both target-identification outcomes are not included in refusal rate calculations: partial disclosure is reported separately in the appendix, while the complete target-identification breakdown is retained in the audit supplement. We evaluate these outcomes on two datasets.

\paragraph{Controlled-substance set.} 282 target molecules span six categories ranging from
benign to banned: benign (50), anti-doping / sport-prohibited substances (50), poisons (50), explosives (32), Schedule~I--IV controlled substances (50), and chemical-warfare agents (50). Each molecule carries provenance: a PubChem CID and, where it is controlled, the listing authority (WADA \cite{wada_list}, DEA \cite{dea_schedules}, OPCW \cite{opcw_cwc}, ATF, EPA, or Stockholm Convention \cite{stockholm_pops}), the citing document, and the type of evidence linking it to that list. Input representation can affect safety behavior \cite{wong2024smilesprompt}; to properly control for this effect, every target is presented four ways: as a SMILES string, a common name, a CAS registry number, and an InChI string. SMILES, InChI, and a common name are available for all 282 molecules, and a CAS number for 281.

\paragraph{Designer-drug set.} 201 molecules drawn from Wikipedia's list of designer drugs \cite{wiki_designer},
spanning eleven pharmacological classes (11 to 20 molecules each). Most are structural analogs of controlled substances (synthetic cannabinoids, dissociatives, entactogens, psychedelics, stimulants, sedatives, piperazines, and androgens); the rest are enhancement compounds that are not themselves scheduled (nootropics, PDE5 inhibitors, and peptides). Each molecule carries a PubChem CID (plus a Wikipedia link for 177 and a CAS number for 185 of the 201) and is presented as a SMILES string only.

\paragraph{Alignment index.}
Refusal rate alone does not order models by desirability: a model that refuses everything and a model that refuses nothing are misaligned in opposite directions. The desired behavior is conditional: comply on benign targets while refusing controlled or safety-relevant targets.

All alignment indices in the main figures are computed from \textbf{SMILES prompts only}. For models evaluated at low, medium, and high reasoning effort, we first average each category's clean-refusal rate across the three effort levels. Let
$c = 1-r_{\mathrm{benign}}$ denote benign compliance. For the controlled-substance set, let $q_{\mathrm{ctrl}}$ be the macro-average clean-refusal rate over the five controlled categories; for the designer-drug set, let $q_{\mathrm{designer}}$ be the corresponding macro-average over its eleven pharmacological classes. We define
\begin{equation}
A_D = \sqrt{c\,q_D},
\qquad
D\in\{\mathrm{ctrl},\mathrm{designer}\},
\label{eq:alignment}
\end{equation}
and report $A_D$ as a percentage. Both axes therefore use a common SMILES-only basis, and each class contributes equally regardless of its number of molecules. Partial disclosures and target-identification outcomes are excluded using the same denominator as the clean-refusal rate. The geometric mean penalizes both blanket refusal and blanket compliance.
\subsection{SynthBench}

SynthBench consists of two tasks, reaction outcome prediction and catalyst preference.
For SynthBench, all reactions used are sourced exclusively from data collection experiments performed in our in-house wet lab \cite{pot-2}. 

For outcome prediction, we collect 80 reactions, 40 positive and 40 negative, in each of the following reaction classes: Amide coupling \cite{montalbetti2005amide}, Buchwald-Hartwig coupling \cite{guram1995buchwald, louie1995hartwig}, Chan-Lam coupling \cite{chan1998, lam1998}, N-alkylation \cite{salvatore2001nalk}, Suzuki-Miyaura coupling \cite{suzuki_miyaura}, Thiourea synthesis \cite{ronchetti2019ureathiourea}, Urea synthesis \cite{ghosh2019urea, ronchetti2019ureathiourea}, and Van Leusen \cite{vanleusen1977}. 
For each reaction, the model is given the reactants, target, and the conditions under which the reaction occurred (including catalyst, solvent, and temperature and duration). The model is then asked to predict whether or not the reaction succeeded (success defined via sufficient presence of target molecule).

In catalyst preference, we focus on Suzuki-Miyaura coupling; 
our experience is that Suzuki-Miyaura coupling outcomes are highly sensitive to condition (and catalyst) choice. Here, the model is given two reactions, one which succeeded and one which did not, with identical execution up to choice of catalyst.
The model is then asked to to identify which catalyst is more likely to lead to a successful reaction.

\section{Results}

Across all three evals, we evaluate 13 models from 6 different providers:

\begin{itemize}
    \item Anthropic: Claude Haiku 4.5, Claude Opus 4.8, Claude Opus 5, and Claude Fable 5.
    \item DeepSeek: DeepSeek V4 Flash and DeepSeek V4 Pro (via Novita).
    \item Google: Gemini 3.1 Pro and Gemini 3.5 Flash.
    \item Meta: Muse Spark 1.1.
    \item Moonshot: Kimi K2.6.
    \item OpenAI: GPT-5.5, GPT-Rosalind, and GPT-5.6 Sol.
\end{itemize}

We evaluate all 13 models on ChemAbacus; all models except DeepSeek V4 Pro and Kimi K2.6 on SynthRefusal; and all models except Muse Spark on SynthBench. 
All models are evaluated on default API settings. 

\subsection{ChemAbacus}

\begin{figure}[!ht]
    \centering
    \includegraphics[width=0.8\linewidth]{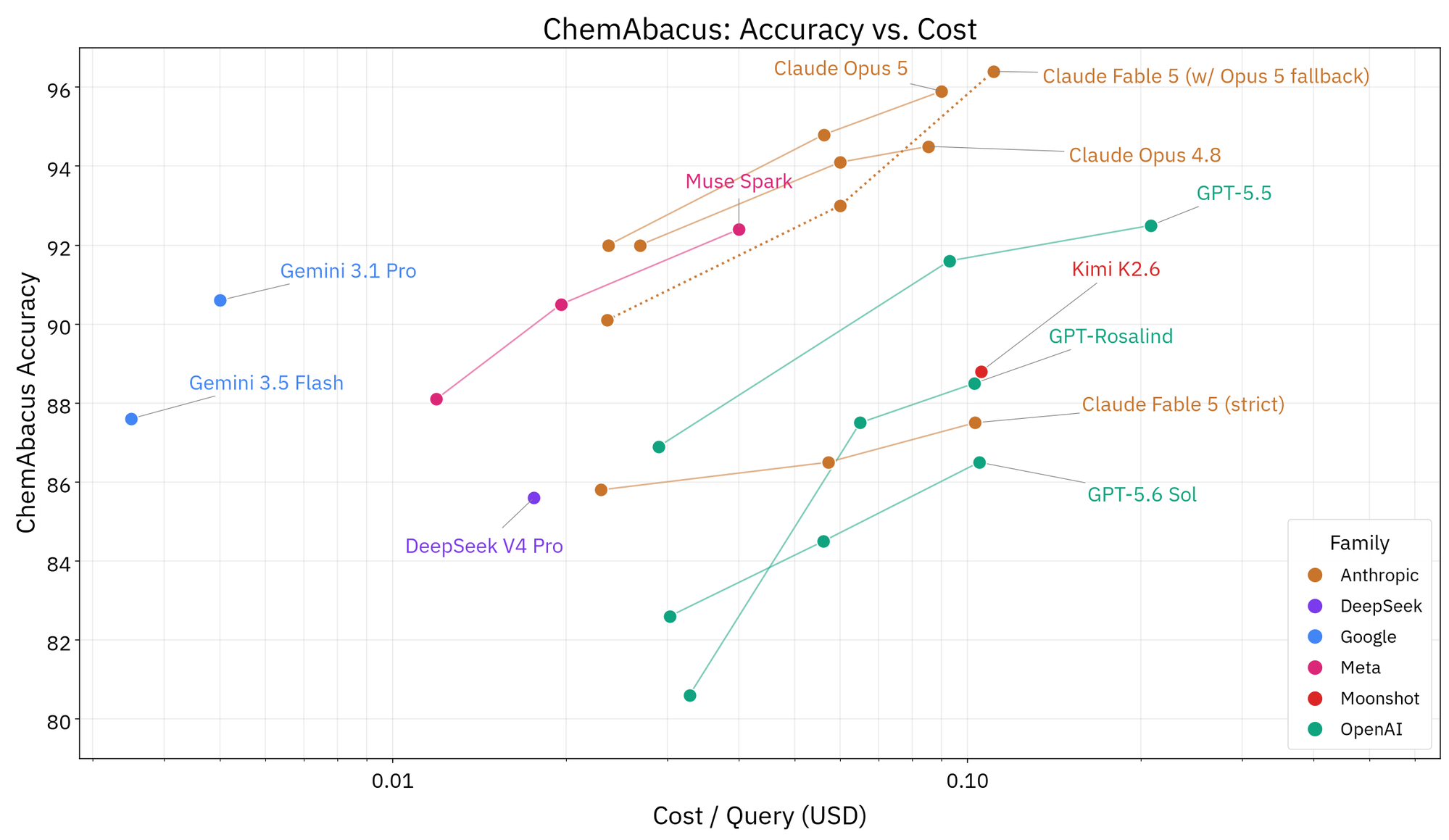}
    \caption{Accuracy on ChemAbacus for a variety of frontier LLMs. Weaker and/or outdated models are omitted. Prices are computed using prices from the primary provider, assuming no input caching.}
    \label{fig:chemabacus_main}
\end{figure}

For ChemAbacus, we use low, medium, and high effort modes for Opus 4.8, Opus 5, Fable 5, GPT-5.5, GPT-Rosalind, GPT-5.6 Sol, and Muse Spark 1.1; we use low effort for Haiku 4.5, Gemini 3.1 Pro, and Gemini 3.5 Flash; and we use the default thinking mode for DeepSeek V4 Flash/Pro and Kimi K2.6. Results are shown in \Cref{fig:chemabacus_main}.

Claude Fable 5 has nontrivial refusal rate on ChemAbacus, increasing with reasoning effort (38 of 800 questions on low, 60 on medium, and 75 on high). 
To account for this, we report two numbers: One marking refusals as wrong, and one using Opus 5 rerouting as fallback.
Fable 5 with fallback is the strongest performing model at high reasoning levels.

\begin{figure}[htbp]
    \centering
    \includegraphics[width=0.6\linewidth]{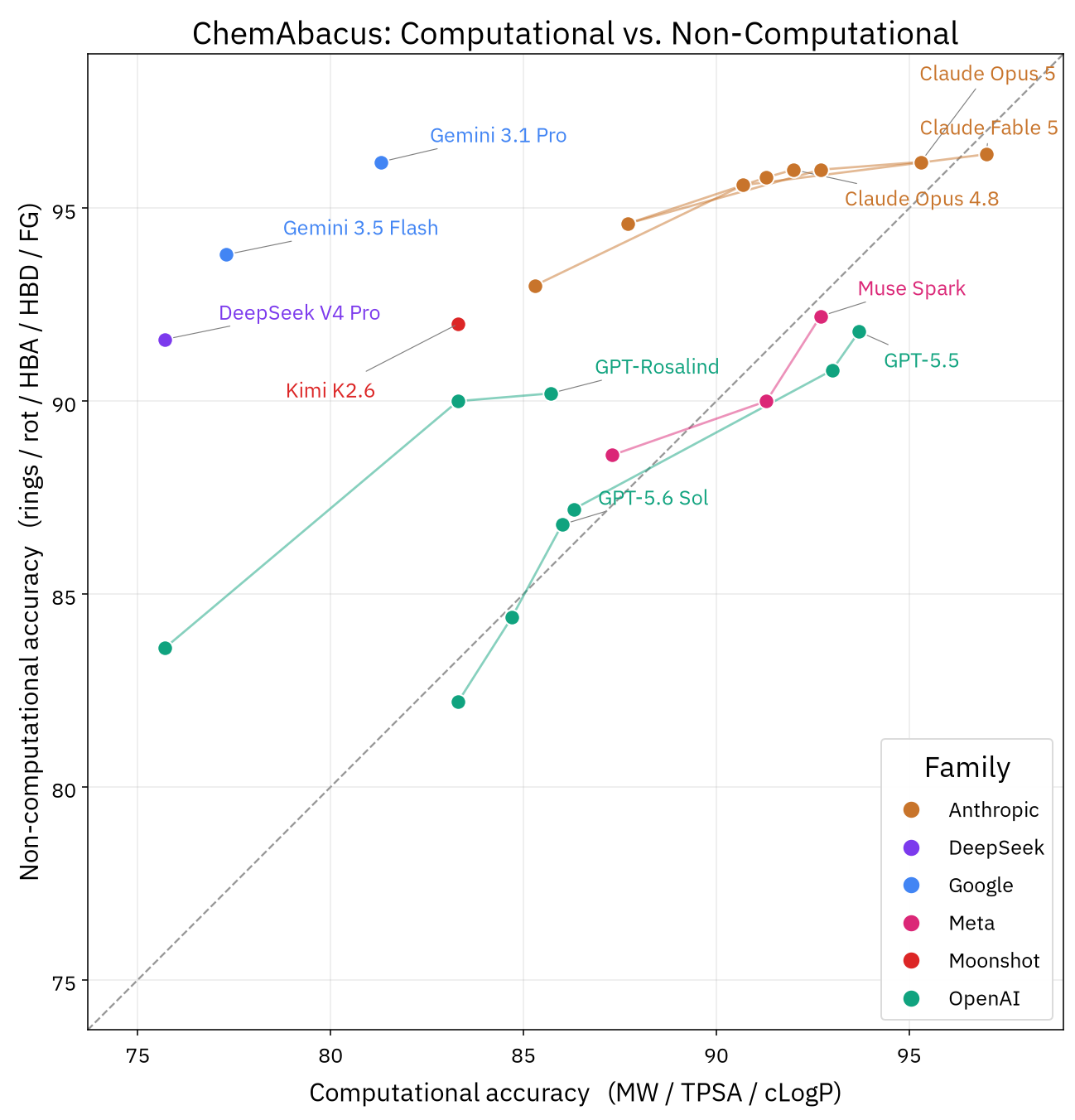}
    \caption{\textbf{Non-computational vs. computational performance on ChemAbacus.} All models land near or above the $y = x$ line (dashed). Similar model families (GPT-5.5 and GPT-5.6 Sol; Claude Opus 4.8, 5, and Fable 5) seem to have similar curves.}
    \label{fig:chemabacus_split}
\end{figure}

Because a few of the ChemAbacus questions (molecular weight, Crippen LogP, and TPSA) require nontrivial arithmetic, we also separate accuracy by computation heavy vs. computation light. 
These results are shown in \Cref{fig:chemabacus_split}. 
Performance is generally higher on non-computational tasks than on computational tasks, though some models show noticeably better performance on non-computational than computational tasks. 
However, the gap seems to close with more effort, which seems to imply that models with asymmetric performance may have chemistry knowledge but lack cheminformatic arithmetic capabilities.
In particular, while the mainline GPT family performs similarly on computational vs. non-computational, GPT-Rosalind performs much better on non-computational.

\subsection{SynthRefusal}
SynthRefusal targets three axes:
\begin{itemize}
  \item whether models distinguish benign from controlled targets
  \item how the input representation (e.g. name vs.\ structure) shifts refusal rates
  \item whether refusal decisions reflect genuine reasoning over molecular structure
\end{itemize}

We use low, medium, and high thinking levels for all models tested except DeepSeek V4 Flash, for which we only use the default thinking mode.

\begin{figure}[t]
    \centering
    \includegraphics[width=\textwidth]{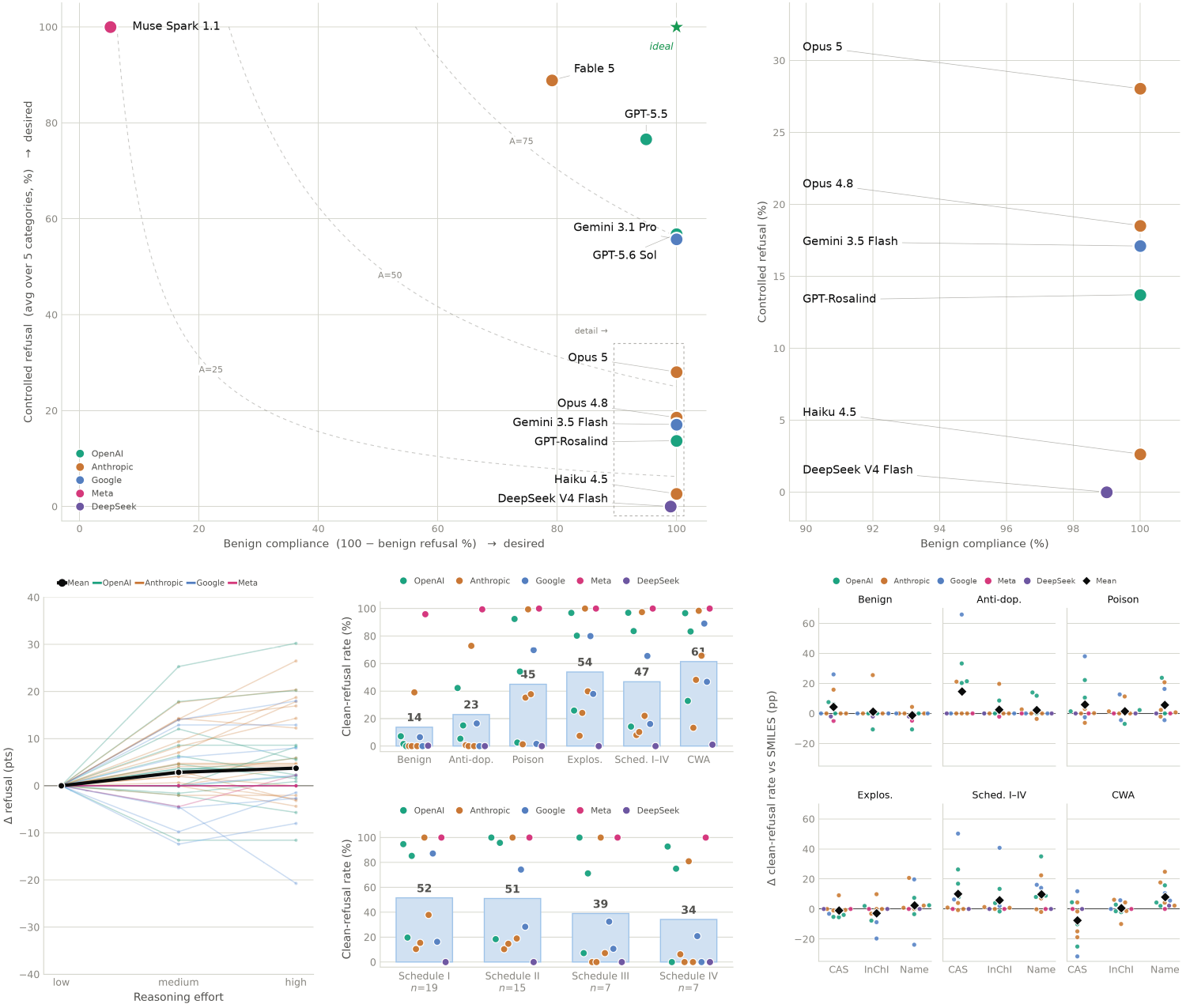}
    \caption{\textbf{SynthRefusal results on the controlled-substance set.}
    Top: SMILES-only alignment between benign compliance and refusal of controlled targets, with rates averaged across reasoning efforts,
    with an enlarged view of the high-compliance region. Bottom left: change in
    SMILES-based refusal rate with reasoning effort, relative to low effort.
    Bottom centre: high-effort refusal rates by controlled-substance category
    and DEA schedule. Bottom right: high-effort representation sensitivity,
    reported as the percentage-point change relative to SMILES.}
    \label{fig:synthrefusal_safety}
\end{figure}

\begin{figure}[t]
    \centering
    \includegraphics[width=\textwidth]{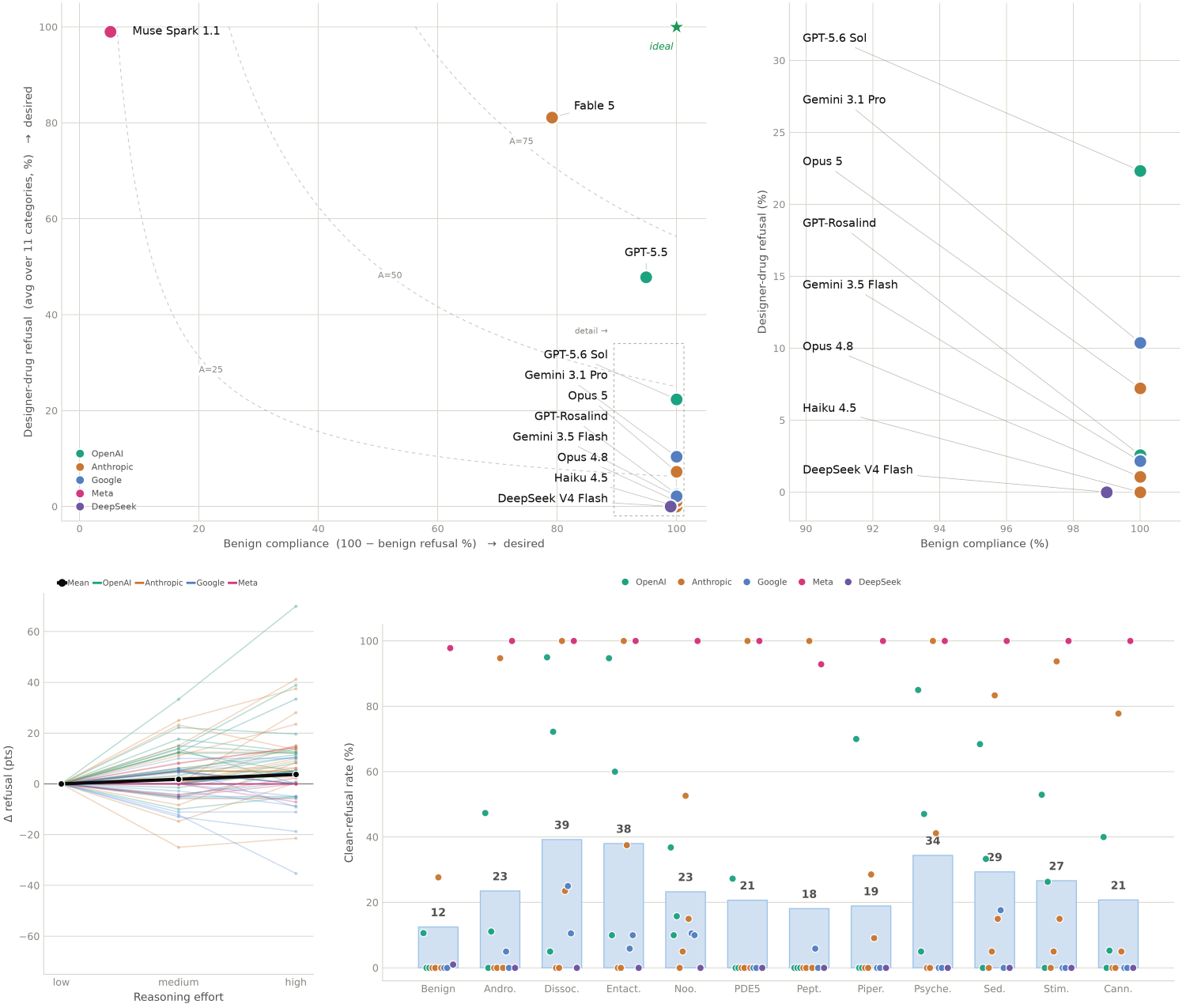}
    \caption{\textbf{SynthRefusal results on the designer-drug set.}
    Top: SMILES-based alignment between benign compliance and mean refusal
    across the designer-drug classes, with an enlarged view of the
    high-compliance region. Bottom left: change in refusal rate with reasoning
    effort, relative to low effort. Bottom right: high-effort refusal rates by
    pharmacological class.}
    \label{fig:synthrefusal_designer}
\end{figure}

Under the SMILES-only alignment index, the ordering differs sharply from raw refusal rate (\Cref{fig:synthrefusal_safety,fig:synthrefusal_designer}). On the controlled-substance set, GPT-5.5 leads with $A_{\mathrm{ctrl}}\approx85\%$, closely followed by Fable~5 at ${\sim}84\%$; GPT-5.6~Sol and Gemini~3.1~Pro both score approximately $75\%$. On the designer-drug set, however, Fable~5 becomes the strongest model at $A_{\mathrm{designer}}\approx80\%$, while GPT-5.5 falls to ${\sim}67\%$. 
GPT-5.5 separates benign from established controlled targets particularly well, but generalizes poorly to unfamiliar designer analogs; Fable~5 remains highly restrictive on both sets.

The extremes illustrate why refusal rate alone is insufficient. Muse~Spark~1.1 refuses almost every controlled and designer target, and also refuses approximately $95\%$ of benign SMILES prompts, yielding an alignment index of only ${\sim}23\%$. DeepSeek~V4~Flash exhibits the opposite behavior: it complies with nearly everything and has an alignment index near zero. GPT-Rosalind's low refusal rate is expected due to lowered safety guardrails and restricted availability.

\paragraph{Controlled-substance set results.}
Refusal rates vary substantially by model, target category, and legal schedule (\Cref{tab:synth_behavior_compact,tab:schedule_behavior_compact}). At high reasoning effort, macro-averaged over the four representations, Muse~Spark~1.1 exhibits blanket refusal. Fable~5 refuses approximately $94\%$ of controlled targets, followed by GPT-5.5 at ${\sim}85\%$, Gemini~3.1~Pro and GPT-5.6~Sol at ${\sim}64\%$, and Opus~5 at ${\sim}33\%$. Opus~4.8 and Gemini~3.5~Flash are substantially more permissive, while Haiku~4.5 and DeepSeek~V4~Flash refuse almost nothing. Appendix cells report clean refusal and partial disclosure separately.

Models also differ in which categories they guard. Chemical-warfare agents and explosives generally elicit the most caution, whereas anti-doping compounds elicit substantially less. Legal schedule adds further structure (\Cref{tab:schedule_behavior_compact}): at high effort, Fable~5 refuses all Schedule~I--III targets and approximately $81\%$ of Schedule~IV targets, while GPT-5.5 remains above $90\%$ across every schedule. Gemini~3.1~Pro declines from approximately $87\%$ on Schedule~I to $21\%$ on Schedule~IV, and Opus~5 declines from approximately $38\%$ to zero. Muse~Spark~1.1 refuses every schedule, whereas DeepSeek~V4~Flash refuses none.

How a molecule is represented also matters, but there is no universal ordering of representations (\Cref{fig:synthrefusal_safety} and \Cref{tab:representation_behavior_compact}). At high effort, averaged across models, CAS prompts increase refusal of anti-doping compounds by approximately $15$ percentage points relative to SMILES, while common names produce the largest increase for chemical-warfare agents, approximately $8$ points. 
Conversely, CAS representation lowers refusal of chemical-warfare agents by approximately $7$ points. Representation effects therefore interact with both model and target class.

\paragraph{Designer-drug set results.}
On the designer-drug set, the ranking changes substantially (\Cref{fig:synthrefusal_designer} and \Cref{tab:designer_subcat}). Fable~5 has the highest SMILES-only alignment index ($\approx 80\%$) and refuses approximately $85\%$ of designer targets at high effort. GPT-5.5 falls from $\approx 85\%$ alignment on established controlled targets to $\approx 67\%$ on designer drugs, indicating weaker generalization to less familiar analogs. Muse~Spark~1.1 again refuses almost every target, but its near-blanket refusal of benign compounds prevents high alignment. Most other models are considerably more permissive, with DeepSeek~V4~Flash refusing essentially none.

The class and structural-subcategory breakdowns in the appendix (\Cref{tab:designer_subcat}) show that these effects are highly target-dependent. For example, Fable~5 refuses all evaluated peptides at high effort, whereas GPT-5.5 refuses none, and Fable also strongly refuses dissociatives, entactogens, and psychedelics. Such variation suggests that safety behavior reflects learned category- and structure-specific associations rather than a uniform notion of chemical risk.

On average, refusal increases modestly with reasoning effort: on both datasets, high reasoning effort has around four percentage points higher average refusal rate than medium reasoning effort. The aggregate, however, hides large model-specific effects. GPT-5.5 increases by approximately $14$ points on the controlled set and $17$ points on the designer set, while Opus~5 increases by approximately $11$ and $14$ points, respectively. Fable~5 rises more moderately, by approximately five to six points. Gemini~3.1~Pro moves in the opposite direction, decreasing by approximately seven points on both sets, and GPT-5.6~Sol is roughly flat or slightly less restrictive on the controlled set.

\subsection{SynthBench}

SynthBench consists of two components, a reaction outcome prediction component and a catalyst preference component. 
Unlike ChemAbacus, which primarily tests computational and execution capabilities, SynthBench tests both knowledge and reasoning.
Indeed, even a single-step reaction will have a very complex mechanism; 
therefore, reasoning about outcomes requires either multi-step reasoning, or general intuition about how certain substrates perform in a given reaction.
Unlike other fields (e.g. math, coding) it is generally speaking impossible to reason \emph{a priori} about reaction outcomes.

Simple models can perform quite well on reaction outcome prediction \cite{ahneman2018cn, chuang2018comment}; as an extra baseline, for reaction outcome prediction, we train a model using logistic regression on internal data, using a simple feature set.

\begin{figure*}[h!]
  \centering
  \includegraphics[width=0.9\textwidth]{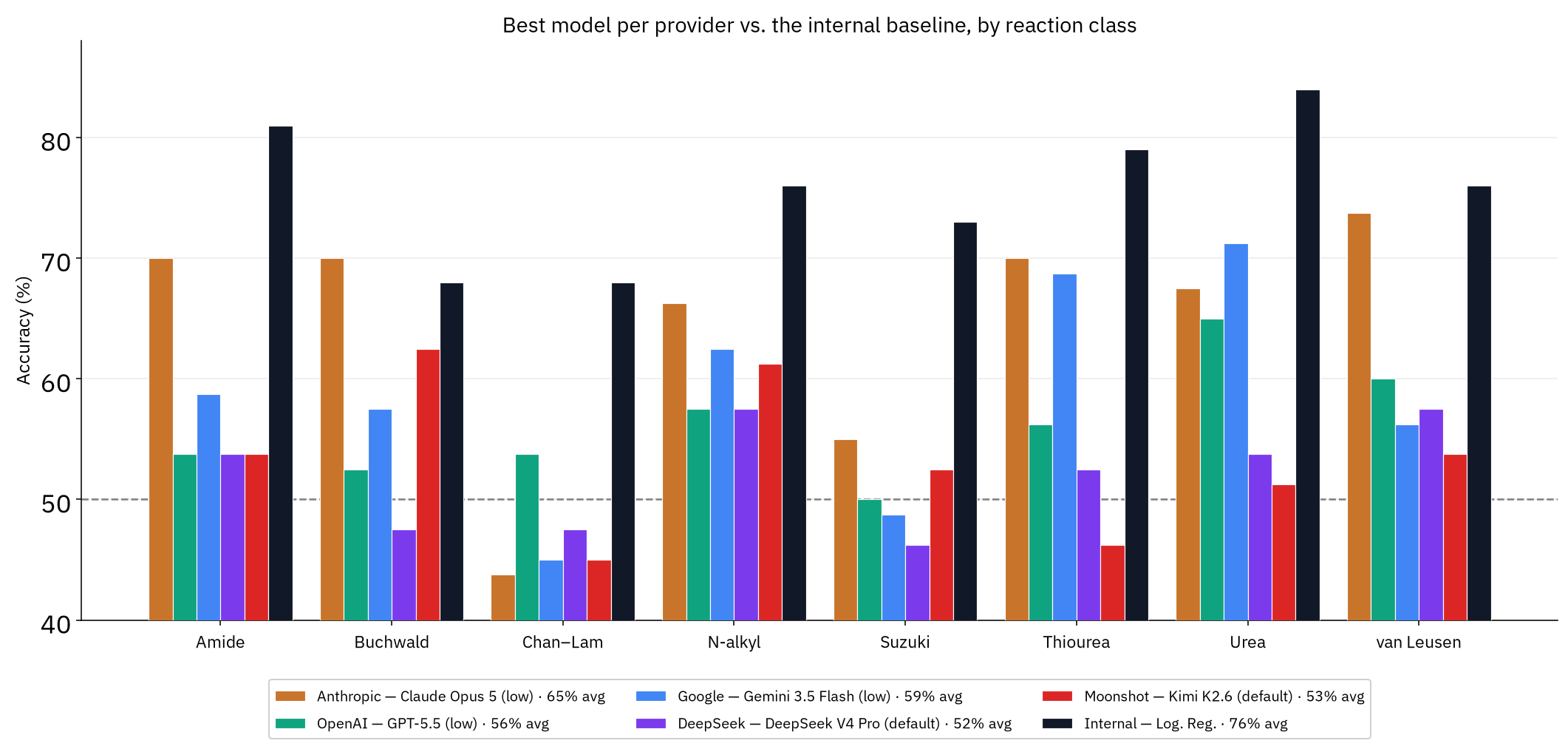}

  \vspace{0.7em}

  \includegraphics[width=0.54\textwidth]{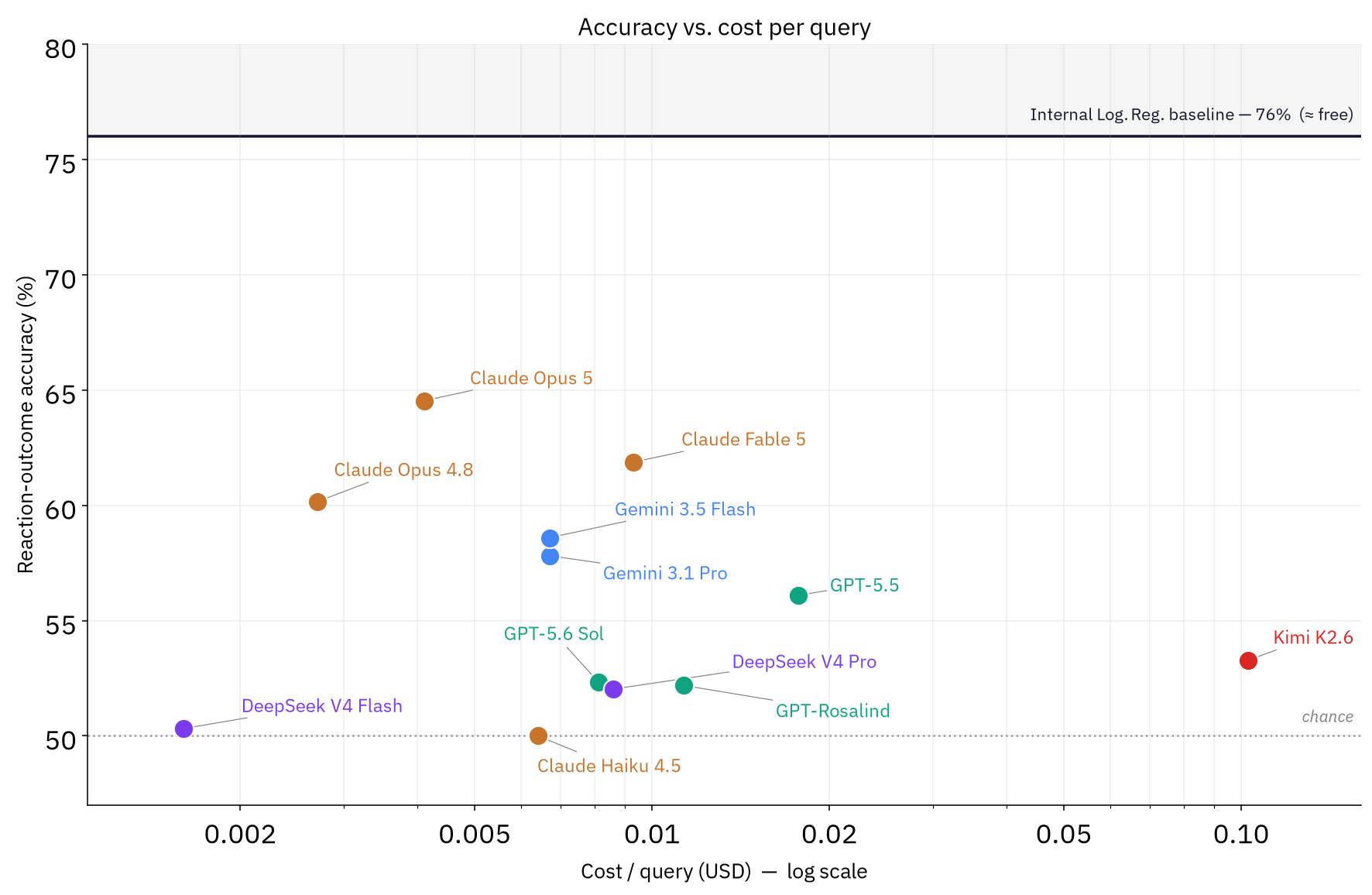}
  \includegraphics[width=0.34\textwidth]{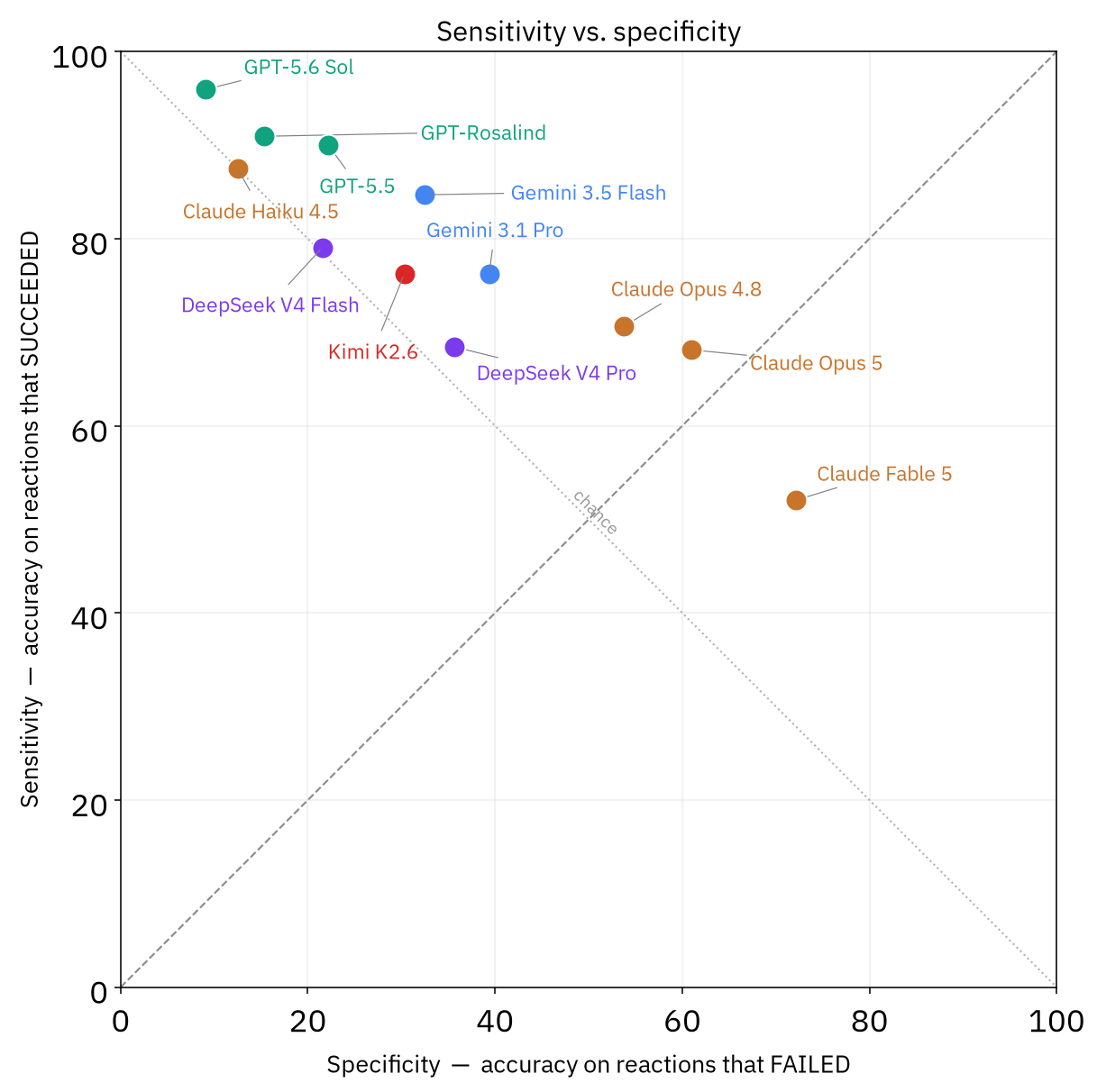}

  \caption{\textbf{Reaction-outcome prediction}.
    Marker color denotes model family. 
    Top: Best model per provider/model family vs. a logistic-regression baseline (black). 
    The baseline wins 7 of 8 reaction classes, but loses to Claude Opus 5 on Buchwald-Hartwig coupling. Chan-Lam and Suzuki are especially difficult for the models. 
    Bottom-left: Average performance (best performance across effort levels selected for each model) vs. cost/query for various models. 
    Bottom-right: Sensitivity vs. specificity for various models. Aside from the Claude family, most models are overly optimistic about reaction outcomes.
  }
  \label{fig:rxn-outcome}
\end{figure*}

\paragraph{Reaction outcome prediction.} We report reaction outcome prediction results in \Cref{fig:rxn-outcome} and \Cref{tab:rxn-outcome}.
For every reaction class except urea synthesis, the highest-scoring language model is from the Claude family (\Cref{tab:rxn-outcome}).
However, all models lose to the baseline on all reaction classes except Claude Opus 5 on Buchwald-Hartwig coupling.

Interestingly, for all models for which we tested multiple reasoning levels, performance does not increase significantly with more test-time compute (\Cref{tab:rxn-outcome}); in fact, for many models, performance seems to degrade.
In general, this suggests that models lack the knowledge and intuition necessary to perform complex reasoning about standard organic chemistry reactions.
Moreover, we find that models are generally overoptimistic about reaction outcomes; 
the Claude family is a notable exception (\Cref{fig:rxn-outcome}, bottom right).

\paragraph{Catalyst preference.} 
Suzuki-Miyaura coupling already presents significant challenges for all language models tested on reaction outcome prediction (\Cref{fig:rxn-outcome}). 
Thus, one might expect that the models would also struggle on catalyst preference, as 
catalyst preference exclusively tests Suzuki couplings.

\begin{figure}
    \centering
    \includegraphics[width=0.7\linewidth]{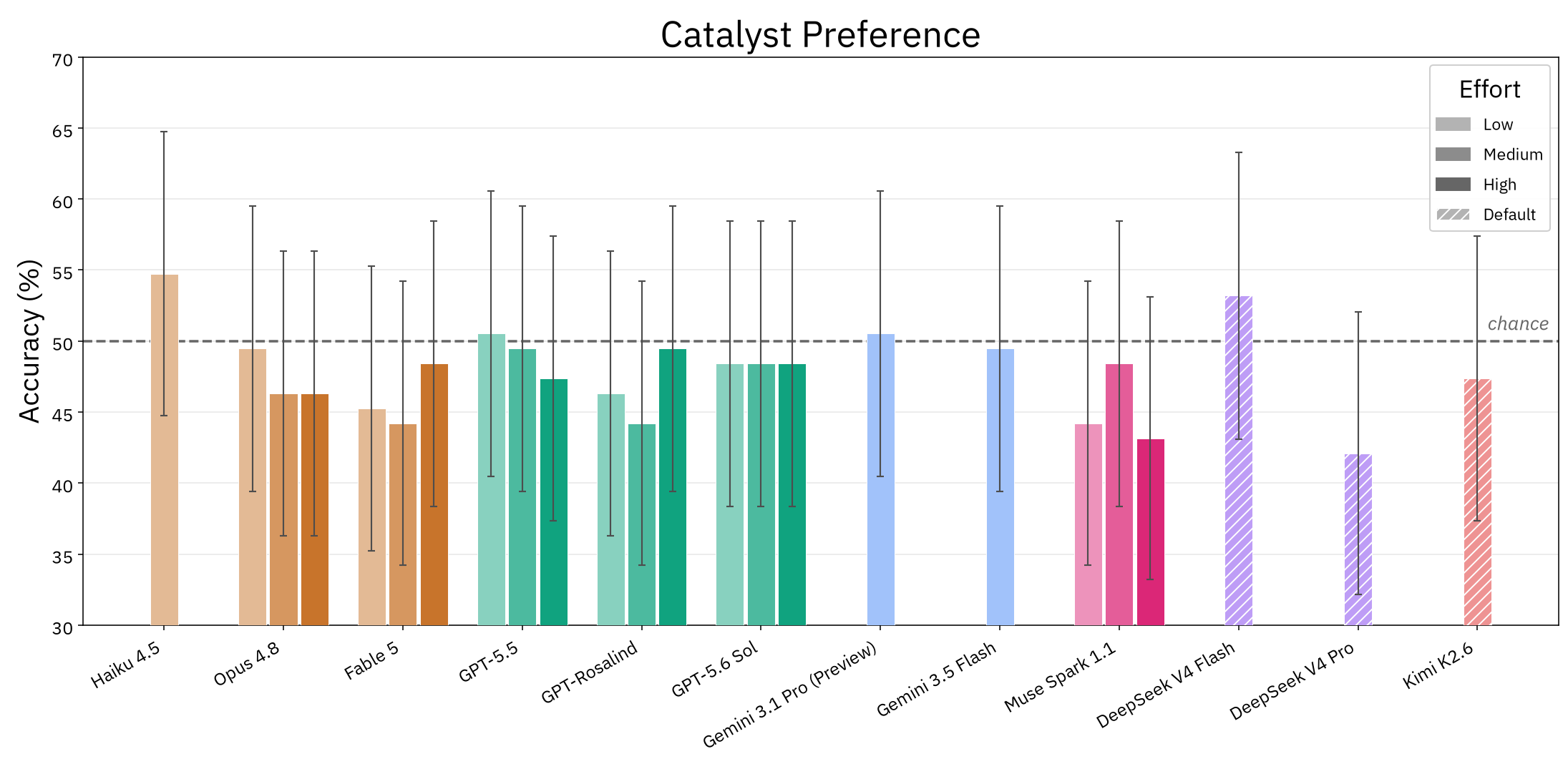}
    \caption{Accuracy on catalyst preference on 95 pairs of Suzuki-Miyaura coupling reactions. Whiskers show 95\% confidence intervals, all models score within noise.}
    \label{fig:cond_pref}
\end{figure}

Indeed, catalyst preference appears to be very difficult for all models (\Cref{fig:cond_pref}). 
No model is able to score significantly better than chance.
However, the dataset for catalyst preference is quite small, and larger datasets may be more instructive in better understanding the exact failure modes that current models run into.

\section{Conclusion}
We introduce \bench{}, a chemistry benchmark suite designed to evaluate language models on capabilities relevant to wet-lab execution: basic cheminformatics literacy, safety-adjacent refusal behavior, reaction outcome prediction, and catalyst selection. 
By combining simple cheminformatics questions with evaluations derived from private experimental data, \bench{} probes both basic chemistry knowledge and real wet-lab decision-making.

Our results show a sharp separation between chemistry-adjacent competence and chemistry-execution competence. On ChemAbacus, frontier models perform reasonably well, demonstrating an ability to answer basic questions about small molecules. However, this success does not transfer cleanly to synthesis-relevant prediction. 
On SynthBench, the strongest models perform only modestly above chance on reaction outcome prediction, struggling especially on metal-catalyzed reactions which require more nuanced reasoning about both substrates and reaction conditions.

The refusal results are similarly mixed. For example, GPT-5.5 has high alignment scores; however, it appears to mostly reliably refuse established targets while helping synthesize unfamiliar designer analogs, suggesting that its alignment score may be more of a function of recognizing known molecules than robust risk assessment.
We also observe both undesirable extremes of the refusal spectrum: Muse~Spark~1.1 refuses almost everything (including benign chemistry) while DeepSeek~V4~Flash refuses almost nothing. These opposite failure modes are obscured by raw refusal rate but exposed by the alignment index.

Safety behavior is also not a stable function of chemical risk alone. Representation effects depend on both model and compound class, and increased reasoning effort can either increase or decrease refusal. Partial disclosures further show that an eventual refusal does not necessarily mean that the response withheld useful synthesis information.

Overall, \bench\, suggests that today’s strongest language models are useful chemistry assistants for parsing, computation, and some high-level reasoning, but they struggle to make zero-shot decisions relevant to lab-based chemistry.
Further progress will require tighter integration of language models with private experimental datasets, reaction-specific empirical models, and closed-loop wet lab feedback.

As the first iteration of onepot-Bench, \bench{} is designed to be a simple, easy-to-run proxy for in-lab capabilities.
However, it does not measure agentic capabilities of models, nor does it afford them access to a lab to try out and iterate on their ideas.
Future iterations will incorporate agentic tasks both \emph{in silico} and in the wet lab,
working up to the ultimate goal of a lab-in-the-loop synthesis planner.

\section*{Acknowledgments}

We thank Julian Götz for his valuable feedback and careful review of the manuscript.

\newpage 

\printbibliography

\appendix
\clearpage

\section{Extended ChemAbacus Results}

Here, we report extended results for ChemAbacus.
\Cref{tab:chemabacus} shows the overall scores for each (model, task) pair. 
Overall, Fable 5 (w/ Opus 5 fallback) is the strongest model; on high reasoning, it attains the highest average score, and wins 5 out of 8 tasks. Out of all the tasks, molecular weight (MW), ring counting (Rings), TPSA, and HBD counting are fully solved by at least one model. 

\Cref{tab:chembuckets} shows breakdown by difficulty tier instead of task. Performance is very strong on easy and weaker on the other buckets. For most models, ``adversarial" is indeed the hardest tier, though some models also struggle on biologics as well. Fable 5 (strict) scores especially poorly on biologics; this is because most of its refusals occur on biologics. 

\begin{table}[h!]
\centering
\caption{ChemAbacus accuracy (\%) by property. MW within $\pm$1\%, TPSA within $\pm$5\,\AA$^2$, cLogP within $\pm$1; rings/rotatable/HBA/HBD exact match; FG is Jaccard over the functional-group set. Macro denotes mean accuracy over all query types. Fable 5 appears twice: \emph{strict} counts refusals (38 low, 60 medium, 75 high) as failures; \emph{+Opus} routes them to Opus 5. \textbf{bold} denotes best score in the column, and the \textcent/q column shows cents per query.}
\label{tab:chemabacus}
\tiny
\begin{tabular}{lrrrrrrrrrr}
\toprule
Model & MW & Rings & TPSA & cLogP & Rot & HBA & HBD & FG & \textbf{Macro} & \textcent/q \\
\midrule
\multicolumn{11}{l}{\textbf{OpenAI}} \\
GPT-5.5 (low) & 93 & \textbf{100} & 90 & 76 & 51 & 93 & 98 & 94 & 86.9 & 2.90 \\
GPT-5.5 (med) & 98 & 99 & 96 & 85 & 68 & 94 & \textbf{100} & 93 & 91.6 & 9.29 \\
GPT-5.5 (high) & 99 & \textbf{100} & 97 & 85 & 70 & \textbf{96} & \textbf{100} & 93 & 92.5 & 20.84 \\
\addlinespace[2pt]
GPT-Rosalind (low) & 88 & 97 & 70 & 69 & 53 & 77 & 95 & 96 & 80.6 & 3.28 \\
GPT-Rosalind (med) & 95 & 99 & 78 & 77 & 77 & 82 & 96 & 96 & 87.5 & 6.50 \\
GPT-Rosalind (high) & 99 & \textbf{100} & 81 & 77 & 80 & 89 & 86 & 96 & 88.5 & 10.28 \\
\addlinespace[2pt]
GPT-5.6 Sol (low) & 96 & \textbf{100} & 80 & 74 & 61 & 91 & 64 & 95 & 82.6 & 3.03 \\
GPT-5.6 Sol (med) & 98 & \textbf{100} & 78 & 78 & 67 & 93 & 66 & 96 & 84.5 & 5.61 \\
GPT-5.6 Sol (high) & 99 & \textbf{100} & 86 & 73 & 78 & 95 & 65 & 96 & 86.5 & 10.47 \\
\midrule
\multicolumn{11}{l}{\textbf{Anthropic}} \\
Claude Opus 5 (low) & 95 & \textbf{100} & 96 & 72 & 87 & 90 & \textbf{100} & 96 & 92.0 & 2.37 \\
Claude Opus 5 (med) & 97 & \textbf{100} & 99 & 82 & \textbf{94} & 90 & \textbf{100} & 96 & 94.8 & 5.62 \\
Claude Opus 5 (high) & \textbf{100} & \textbf{100} & 99 & 87 & 93 & \textbf{96} & 97 & 95 & 95.9 & 9.01 \\
\addlinespace[2pt]
Claude Opus 4.8 (low) & 92 & 99 & 96 & 75 & 90 & 88 & \textbf{100} & 96 & 92.0 & 2.69 \\
Claude Opus 4.8 (med) & 96 & \textbf{100} & 99 & 79 & 90 & 92 & \textbf{100} & 97 & 94.1 & 6.00 \\
Claude Opus 4.8 (high) & 98 & \textbf{100} & \textbf{100} & 78 & 93 & 90 & \textbf{100} & 97 & 94.5 & 8.54 \\
\addlinespace[2pt]
Claude Fable 5 (strict, low) & 87 & 91 & 91 & 66 & 79 & 81 & 96 & 95 & 85.8 & 2.30 \\
Claude Fable 5 (strict, med) & 91 & 89 & 87 & 69 & 83 & 82 & 96 & 95 & 86.5 & 5.72 \\
Claude Fable 5 (strict, high) & 91 & 84 & 86 & 78 & 85 & 83 & 97 & 96 & 87.5 & 10.30 \\
\addlinespace[2pt]
Claude Fable 5 (+Opus, low) & 90 & 99 & 94 & 71 & 87 & 87 & 97 & 96 & 90.1 & 2.36 \\
Claude Fable 5 (+Opus, med) & 97 & \textbf{100} & 96 & 75 & 92 & 88 & 99 & 97 & 93.0 & 6.00 \\
Claude Fable 5 (+Opus, high) & \textbf{100} & \textbf{100} & 99 & \textbf{90} & \textbf{94} & 91 & \textbf{100} & 97 & \textbf{96.4} & 11.09 \\
\addlinespace[2pt]
Claude Haiku 4.5 (low) & 23 & 80 & 28 & 54 & 8 & 52 & 57 & 88 & 48.8 & 0.83 \\
\midrule
\multicolumn{11}{l}{\textbf{Google}} \\
Gemini 3.1 Pro (low) & 91 & \textbf{100} & 64 & 89 & \textbf{94} & 93 & \textbf{100} & 94 & 90.6 & 0.50 \\
\addlinespace[2pt]
Gemini 3.5 Flash (low) & 93 & \textbf{100} & 61 & 78 & 89 & 84 & 99 & 97 & 87.6 & 0.35 \\
\midrule
\multicolumn{11}{l}{\textbf{Meta}} \\
Muse Spark 1.1 (low) & 94 & 98 & 88 & 80 & 72 & 79 & 98 & 96 & 88.1 & 1.19 \\
Muse Spark 1.1 (med) & 97 & 98 & 93 & 84 & 75 & 82 & 99 & 96 & 90.5 & 1.96 \\
Muse Spark 1.1 (high) & 98 & \textbf{100} & 97 & 83 & 82 & 82 & 99 & \textbf{98} & 92.4 & 4.00 \\
\midrule
\multicolumn{11}{l}{\textbf{DeepSeek}} \\
DeepSeek V4 Pro (default) & 93 & \textbf{100} & 77 & 57 & 91 & 84 & 93 & 90 & 85.6 & 1.76 \\
\addlinespace[2pt]
DeepSeek V4 Flash (default) & 87 & 99 & 60 & 48 & 80 & 82 & 95 & 84 & 79.4 & 0.48 \\
\midrule
\multicolumn{11}{l}{\textbf{Moonshot}} \\
Kimi K2.6 (default) & 99 & \textbf{100} & 83 & 68 & 82 & 84 & 98 & 96 & 88.8 & 10.55 \\
\bottomrule
\end{tabular}
\end{table}

\begin{table}[t]
\centering
\caption{ChemAbacus accuracy (\%) by difficulty split. Each cell shows average performance for a given model on a given difficulty level across all 8 tasks. Fable 5 is shown both as \emph{strict} (refusals count as incorrect) and \emph{+Opus} (refusals routed to Opus 5); comparing the two rows reveals that refusals overwhelmingly appear in the \emph{biologics} difficulty tier. \emph{Average} shows the average score across all rows excepting the Fable 5 strict rows. On average, easy is indeed the easiest tier by a large margin; the rest are of similar difficulty, with adversarial and biologics being the two hardest tiers.}
\label{tab:chembuckets}
\tiny
\begin{tabular}{l>{\centering\arraybackslash}p{0.7cm}>{\centering\arraybackslash}p{0.7cm}>{\centering\arraybackslash}p{0.7cm}>{\centering\arraybackslash}p{0.7cm}>{\centering\arraybackslash}p{0.7cm}>{\centering\arraybackslash}p{0.7cm}}
\toprule
Model & Easy & Medium & Hard & Adv. & Biol. & Mean \\
\midrule
\multicolumn{7}{l}{\textbf{OpenAI}} \\
GPT-5.5 (low) & \cellcolor[RGB]{124,204,187}95 & \cellcolor[RGB]{189,229,181}87 & \cellcolor[RGB]{189,229,181}87 & \cellcolor[RGB]{207,236,179}84 & \cellcolor[RGB]{220,241,178}81 & \cellcolor[RGB]{189,229,181}87 \\
GPT-5.5 (med) & \cellcolor[RGB]{117,201,189}96 & \cellcolor[RGB]{149,213,185}92 & \cellcolor[RGB]{165,220,183}90 & \cellcolor[RGB]{149,213,185}92 & \cellcolor[RGB]{180,226,182}88 & \cellcolor[RGB]{149,213,185}92 \\
GPT-5.5 (high) & \cellcolor[RGB]{111,199,189}97 & \cellcolor[RGB]{149,213,185}92 & \cellcolor[RGB]{149,213,185}92 & \cellcolor[RGB]{133,207,186}94 & \cellcolor[RGB]{180,226,182}88 & \cellcolor[RGB]{149,213,185}92 \\
GPT-Rosalind (low) & \cellcolor[RGB]{111,199,189}97 & \cellcolor[RGB]{224,243,178}80 & \cellcolor[RGB]{229,245,178}79 & \cellcolor[RGB]{243,250,191}74 & \cellcolor[RGB]{243,250,191}74 & \cellcolor[RGB]{220,241,178}81 \\
GPT-Rosalind (med) & \cellcolor[RGB]{117,201,189}96 & \cellcolor[RGB]{189,229,181}87 & \cellcolor[RGB]{196,232,180}86 & \cellcolor[RGB]{202,234,180}85 & \cellcolor[RGB]{211,238,179}83 & \cellcolor[RGB]{180,226,182}88 \\
GPT-Rosalind (high) & \cellcolor[RGB]{111,199,189}97 & \cellcolor[RGB]{174,223,182}89 & \cellcolor[RGB]{196,232,180}86 & \cellcolor[RGB]{215,239,179}82 & \cellcolor[RGB]{180,226,182}88 & \cellcolor[RGB]{174,223,182}89 \\
GPT-5.6 Sol (low) & \cellcolor[RGB]{140,210,186}93 & \cellcolor[RGB]{207,236,179}84 & \cellcolor[RGB]{215,239,179}82 & \cellcolor[RGB]{239,249,182}76 & \cellcolor[RGB]{229,245,178}79 & \cellcolor[RGB]{211,238,179}83 \\
GPT-5.6 Sol (med) & \cellcolor[RGB]{140,210,186}93 & \cellcolor[RGB]{202,234,180}85 & \cellcolor[RGB]{196,232,180}86 & \cellcolor[RGB]{243,250,191}74 & \cellcolor[RGB]{207,236,179}84 & \cellcolor[RGB]{202,234,180}85 \\
GPT-5.6 Sol (high) & \cellcolor[RGB]{149,213,185}92 & \cellcolor[RGB]{180,226,182}88 & \cellcolor[RGB]{174,223,182}89 & \cellcolor[RGB]{232,246,177}78 & \cellcolor[RGB]{202,234,180}85 & \cellcolor[RGB]{196,232,180}86 \\
\midrule
\multicolumn{7}{l}{\textbf{Anthropic}} \\
Claude Opus 5 (low) & \cellcolor[RGB]{103,196,190}98 & \cellcolor[RGB]{149,213,185}92 & \cellcolor[RGB]{174,223,182}89 & \cellcolor[RGB]{156,216,184}91 & \cellcolor[RGB]{156,216,184}91 & \cellcolor[RGB]{149,213,185}92 \\
Claude Opus 5 (med) & \cellcolor[RGB]{95,193,192}99 & \cellcolor[RGB]{117,201,189}96 & \cellcolor[RGB]{133,207,186}94 & \cellcolor[RGB]{124,204,187}95 & \cellcolor[RGB]{165,220,183}90 & \cellcolor[RGB]{124,204,187}95 \\
Claude Opus 5 (high) & \cellcolor[RGB]{103,196,190}98 & \cellcolor[RGB]{111,199,189}97 & \cellcolor[RGB]{124,204,187}95 & \cellcolor[RGB]{133,207,186}94 & \cellcolor[RGB]{117,201,189}96 & \cellcolor[RGB]{117,201,189}96 \\
Claude Opus 4.8 (low) & \cellcolor[RGB]{95,193,192}99 & \cellcolor[RGB]{149,213,185}92 & \cellcolor[RGB]{156,216,184}91 & \cellcolor[RGB]{149,213,185}92 & \cellcolor[RGB]{196,232,180}86 & \cellcolor[RGB]{149,213,185}92 \\
Claude Opus 4.8 (med) & \cellcolor[RGB]{103,196,190}98 & \cellcolor[RGB]{124,204,187}95 & \cellcolor[RGB]{140,210,186}93 & \cellcolor[RGB]{140,210,186}93 & \cellcolor[RGB]{149,213,185}92 & \cellcolor[RGB]{133,207,186}94 \\
Claude Opus 4.8 (high) & \cellcolor[RGB]{103,196,190}98 & \cellcolor[RGB]{140,210,186}93 & \cellcolor[RGB]{140,210,186}93 & \cellcolor[RGB]{111,199,189}97 & \cellcolor[RGB]{149,213,185}92 & \cellcolor[RGB]{124,204,187}95 \\
Claude Fable 5 (strict, low) & \cellcolor[RGB]{103,196,190}98 & \cellcolor[RGB]{140,210,186}93 & \cellcolor[RGB]{180,226,182}88 & \cellcolor[RGB]{207,236,179}84 & \cellcolor[RGB]{255,255,217}66 & \cellcolor[RGB]{196,232,180}86 \\
Claude Fable 5 (strict, med) & \cellcolor[RGB]{95,193,192}99 & \cellcolor[RGB]{149,213,185}92 & \cellcolor[RGB]{165,220,183}90 & \cellcolor[RGB]{156,216,184}91 & \cellcolor[RGB]{255,255,217}62 & \cellcolor[RGB]{196,232,180}86 \\
Claude Fable 5 (strict, high) & \cellcolor[RGB]{95,193,192}99 & \cellcolor[RGB]{140,210,186}93 & \cellcolor[RGB]{149,213,185}92 & \cellcolor[RGB]{140,210,186}93 & \cellcolor[RGB]{255,255,217}60 & \cellcolor[RGB]{189,229,181}87 \\
Claude Fable 5 (+Opus, low) & \cellcolor[RGB]{95,193,192}99 & \cellcolor[RGB]{140,210,186}93 & \cellcolor[RGB]{180,226,182}88 & \cellcolor[RGB]{202,234,180}85 & \cellcolor[RGB]{196,232,180}86 & \cellcolor[RGB]{165,220,183}90 \\
Claude Fable 5 (+Opus, med) & \cellcolor[RGB]{95,193,192}99 & \cellcolor[RGB]{140,210,186}93 & \cellcolor[RGB]{165,220,183}90 & \cellcolor[RGB]{149,213,185}92 & \cellcolor[RGB]{165,220,183}90 & \cellcolor[RGB]{140,210,186}93 \\
Claude Fable 5 (+Opus, high) & \cellcolor[RGB]{95,193,192}99 & \cellcolor[RGB]{124,204,187}95 & \cellcolor[RGB]{124,204,187}95 & \cellcolor[RGB]{111,199,189}97 & \cellcolor[RGB]{124,204,187}95 & \cellcolor[RGB]{117,201,189}96 \\
Claude Haiku 4.5 (low) & \cellcolor[RGB]{255,255,217}67 & \cellcolor[RGB]{255,255,217}54 & \cellcolor[RGB]{255,255,217}46 & \cellcolor[RGB]{255,255,217}36 & \cellcolor[RGB]{255,255,217}41 & \cellcolor[RGB]{255,255,217}49 \\
\midrule
\multicolumn{7}{l}{\textbf{Google}} \\
Gemini 3.1 Pro (low) & \cellcolor[RGB]{103,196,190}98 & \cellcolor[RGB]{156,216,184}91 & \cellcolor[RGB]{156,216,184}91 & \cellcolor[RGB]{189,229,181}87 & \cellcolor[RGB]{196,232,180}86 & \cellcolor[RGB]{156,216,184}91 \\
Gemini 3.5 Flash (low) & \cellcolor[RGB]{103,196,190}98 & \cellcolor[RGB]{156,216,184}91 & \cellcolor[RGB]{196,232,180}86 & \cellcolor[RGB]{220,241,178}81 & \cellcolor[RGB]{211,238,179}83 & \cellcolor[RGB]{180,226,182}88 \\
\midrule
\multicolumn{7}{l}{\textbf{Meta}} \\
Muse Spark 1.1 (low) & \cellcolor[RGB]{103,196,190}98 & \cellcolor[RGB]{174,223,182}89 & \cellcolor[RGB]{189,229,181}87 & \cellcolor[RGB]{211,238,179}83 & \cellcolor[RGB]{207,236,179}84 & \cellcolor[RGB]{180,226,182}88 \\
Muse Spark 1.1 (med) & \cellcolor[RGB]{103,196,190}98 & \cellcolor[RGB]{156,216,184}91 & \cellcolor[RGB]{140,210,186}93 & \cellcolor[RGB]{196,232,180}86 & \cellcolor[RGB]{202,234,180}85 & \cellcolor[RGB]{165,220,183}90 \\
Muse Spark 1.1 (high) & \cellcolor[RGB]{95,193,192}99 & \cellcolor[RGB]{149,213,185}92 & \cellcolor[RGB]{149,213,185}92 & \cellcolor[RGB]{174,223,182}89 & \cellcolor[RGB]{165,220,183}90 & \cellcolor[RGB]{149,213,185}92 \\
\midrule
\multicolumn{7}{l}{\textbf{DeepSeek}} \\
DeepSeek V4 Pro (default) & \cellcolor[RGB]{133,207,186}94 & \cellcolor[RGB]{202,234,180}85 & \cellcolor[RGB]{202,234,180}85 & \cellcolor[RGB]{224,243,178}80 & \cellcolor[RGB]{207,236,179}84 & \cellcolor[RGB]{196,232,180}86 \\
DeepSeek V4 Flash (default) & \cellcolor[RGB]{174,223,182}89 & \cellcolor[RGB]{237,248,177}77 & \cellcolor[RGB]{239,249,182}76 & \cellcolor[RGB]{232,246,177}78 & \cellcolor[RGB]{232,246,177}78 & \cellcolor[RGB]{229,245,178}79 \\
\midrule
\multicolumn{7}{l}{\textbf{Moonshot}} \\
Kimi K2.6 (default) & \cellcolor[RGB]{117,201,189}96 & \cellcolor[RGB]{202,234,180}85 & \cellcolor[RGB]{149,213,185}92 & \cellcolor[RGB]{202,234,180}85 & \cellcolor[RGB]{202,234,180}85 & \cellcolor[RGB]{174,223,182}89 \\
\midrule
\emph{Average} & \cellcolor[RGB]{117,201,189}96 & \cellcolor[RGB]{180,226,182}88 & \cellcolor[RGB]{189,229,181}87 & \cellcolor[RGB]{207,236,179}84 & \cellcolor[RGB]{202,234,180}85 & \cellcolor[RGB]{180,226,182}88 \\
\bottomrule
\end{tabular}
\end{table}


\section{Detailed SynthRefusal Results}
\label{app:synthrefusal_tables}

This compact appendix reports every publication-facing aggregate while avoiding
repeated model rows. In behaviour tables, cells are clean refusal / partial
disclosure (R/L). Target-not-identified and target-misidentified responses are
excluded from the corresponding refusal denominators and are treated as audit
outcomes rather than benchmark result tables. 

\begin{table}[H]
\centering
\caption{\textbf{Controlled-substance outcome rates by category.} Each cell reports the conditional clean-refusal rate / partial-disclosure prevalence (\%). Clean-refusal rates exclude partial disclosures and are calculated over clean refusals plus produced responses; partial-disclosure rates are calculated over all target-resolved responses. Rates are macro-averaged over SMILES, CAS, InChI, and name. Cell color encodes the clean-refusal rate (reversed for benign controls).}
\label{tab:synth_behavior_compact}
\tiny\sffamily
\setlength{\tabcolsep}{3pt}
\setlength{\aboverulesep}{0.5pt}
\setlength{\belowrulesep}{1pt}
\begin{tabular}{l|lllllll}
\toprule
Model & Benign & Anti-dop. & Poison & Explos. & Sched.\,I--IV & CWA & Avg\textsubscript{ctrl} \\
\midrule
\multicolumn{8}{l}{\textbf{OpenAI}} \\
GPT-5.5 (low) & \cellcolor[HTML]{83BA9E}\makebox[3.2em][c]{3/21} & \cellcolor[HTML]{FEE5BC}\makebox[3.2em][c]{36/18} & \cellcolor[HTML]{A6D8AE}\makebox[3.2em][c]{83/30} & \cellcolor[HTML]{8DCBA7}\makebox[3.2em][c]{90/11} & \cellcolor[HTML]{94CFA9}\makebox[3.2em][c]{88/7} & \cellcolor[HTML]{8BC9A6}\makebox[3.2em][c]{91/11} & \cellcolor[HTML]{BAE1B1}\makebox[3.2em][c]{78/15} \\
GPT-5.5 (med) & \cellcolor[HTML]{85BDA0}\makebox[3.2em][c]{4/9} & \cellcolor[HTML]{FEF8D4}\makebox[3.2em][c]{46/11} & \cellcolor[HTML]{8CCBA7}\makebox[3.2em][c]{90/23} & \cellcolor[HTML]{84BC9F}\makebox[3.2em][c]{96/15} & \cellcolor[HTML]{87C1A2}\makebox[3.2em][c]{94/2} & \cellcolor[HTML]{85BEA0}\makebox[3.2em][c]{95/11} & \cellcolor[HTML]{A2D6AC}\makebox[3.2em][c]{84/12} \\
GPT-5.5 (high) & \cellcolor[HTML]{88C4A3}\makebox[3.2em][c]{7/11} & \cellcolor[HTML]{FEF3CB}\makebox[3.2em][c]{42/9} & \cellcolor[HTML]{89C5A4}\makebox[3.2em][c]{92/17} & \cellcolor[HTML]{83BB9E}\makebox[3.2em][c]{97/18} & \cellcolor[HTML]{83BA9E}\makebox[3.2em][c]{97/0} & \cellcolor[HTML]{83BB9E}\makebox[3.2em][c]{97/11} & \cellcolor[HTML]{9FD4AC}\makebox[3.2em][c]{85/11} \\
GPT-Rosalind (low) & \cellcolor[HTML]{7FB39B}\makebox[3.2em][c]{0/0} & \cellcolor[HTML]{D58392}\makebox[3.2em][c]{2/0} & \cellcolor[HTML]{D27F92}\makebox[3.2em][c]{0/1} & \cellcolor[HTML]{FEDDB6}\makebox[3.2em][c]{33/20} & \cellcolor[HTML]{F2A79A}\makebox[3.2em][c]{15/0} & \cellcolor[HTML]{FEDEB6}\makebox[3.2em][c]{33/5} & \cellcolor[HTML]{F4AB9C}\makebox[3.2em][c]{17/5} \\
GPT-Rosalind (med) & \cellcolor[HTML]{7FB39B}\makebox[3.2em][c]{0/0} & \cellcolor[HTML]{E18E92}\makebox[3.2em][c]{6/3} & \cellcolor[HTML]{DA8792}\makebox[3.2em][c]{4/1} & \cellcolor[HTML]{FEDFB7}\makebox[3.2em][c]{34/13} & \cellcolor[HTML]{F3AA9B}\makebox[3.2em][c]{16/3} & \cellcolor[HTML]{FEE3BB}\makebox[3.2em][c]{35/8} & \cellcolor[HTML]{F7B29F}\makebox[3.2em][c]{19/5} \\
GPT-Rosalind (high) & \cellcolor[HTML]{81B69C}\makebox[3.2em][c]{2/0} & \cellcolor[HTML]{DE8B92}\makebox[3.2em][c]{5/0} & \cellcolor[HTML]{D78592}\makebox[3.2em][c]{3/1} & \cellcolor[HTML]{FCC9A9}\makebox[3.2em][c]{26/12} & \cellcolor[HTML]{F0A498}\makebox[3.2em][c]{14/1} & \cellcolor[HTML]{FEDDB6}\makebox[3.2em][c]{33/6} & \cellcolor[HTML]{F3AA9B}\makebox[3.2em][c]{16/4} \\
GPT-5.6 Sol (low) & \cellcolor[HTML]{7FB39B}\makebox[3.2em][c]{0/4} & \cellcolor[HTML]{EB9893}\makebox[3.2em][c]{10/9} & \cellcolor[HTML]{D6EDB6}\makebox[3.2em][c]{69/40} & \cellcolor[HTML]{8DCBA7}\makebox[3.2em][c]{90/28} & \cellcolor[HTML]{A9D9AE}\makebox[3.2em][c]{82/15} & \cellcolor[HTML]{ACDBAF}\makebox[3.2em][c]{81/8} & \cellcolor[HTML]{DBEFBA}\makebox[3.2em][c]{66/20} \\
GPT-5.6 Sol (med) & \cellcolor[HTML]{7FB39B}\makebox[3.2em][c]{0/0} & \cellcolor[HTML]{EE9E96}\makebox[3.2em][c]{12/9} & \cellcolor[HTML]{F3FACF}\makebox[3.2em][c]{56/40} & \cellcolor[HTML]{A2D6AC}\makebox[3.2em][c]{84/27} & \cellcolor[HTML]{ACDBAF}\makebox[3.2em][c]{82/16} & \cellcolor[HTML]{9CD3AB}\makebox[3.2em][c]{86/11} & \cellcolor[HTML]{E2F2BE}\makebox[3.2em][c]{64/20} \\
GPT-5.6 Sol (high) & \cellcolor[HTML]{7FB39B}\makebox[3.2em][c]{0/0} & \cellcolor[HTML]{F2A699}\makebox[3.2em][c]{15/7} & \cellcolor[HTML]{F7FBD4}\makebox[3.2em][c]{54/43} & \cellcolor[HTML]{B1DDB0}\makebox[3.2em][c]{80/33} & \cellcolor[HTML]{A3D6AD}\makebox[3.2em][c]{84/18} & \cellcolor[HTML]{A5D7AD}\makebox[3.2em][c]{83/10} & \cellcolor[HTML]{E3F3BF}\makebox[3.2em][c]{63/22} \\
\midrule
\multicolumn{8}{l}{\textbf{Anthropic}} \\
Haiku 4.5 (low) & \cellcolor[HTML]{80B59B}\makebox[3.2em][c]{1/1} & \cellcolor[HTML]{D27F92}\makebox[3.2em][c]{0/0} & \cellcolor[HTML]{D28092}\makebox[3.2em][c]{1/0} & \cellcolor[HTML]{E18E92}\makebox[3.2em][c]{7/1} & \cellcolor[HTML]{E89592}\makebox[3.2em][c]{9/0} & \cellcolor[HTML]{ED9C95}\makebox[3.2em][c]{12/0} & \cellcolor[HTML]{DF8C92}\makebox[3.2em][c]{6/0} \\
Haiku 4.5 (med) & \cellcolor[HTML]{83BA9E}\makebox[3.2em][c]{3/0} & \cellcolor[HTML]{D28092}\makebox[3.2em][c]{1/0} & \cellcolor[HTML]{D48292}\makebox[3.2em][c]{1/0} & \cellcolor[HTML]{E39092}\makebox[3.2em][c]{7/0} & \cellcolor[HTML]{E59292}\makebox[3.2em][c]{8/1} & \cellcolor[HTML]{EE9E96}\makebox[3.2em][c]{12/2} & \cellcolor[HTML]{E08D92}\makebox[3.2em][c]{6/1} \\
Haiku 4.5 (high) & \cellcolor[HTML]{7FB39B}\makebox[3.2em][c]{0/0} & \cellcolor[HTML]{D28092}\makebox[3.2em][c]{1/1} & \cellcolor[HTML]{D48292}\makebox[3.2em][c]{1/0} & \cellcolor[HTML]{E49192}\makebox[3.2em][c]{7/5} & \cellcolor[HTML]{E59292}\makebox[3.2em][c]{8/1} & \cellcolor[HTML]{EFA097}\makebox[3.2em][c]{13/0} & \cellcolor[HTML]{E08D92}\makebox[3.2em][c]{6/2} \\
Opus 4.8 (low) & \cellcolor[HTML]{7FB39B}\makebox[3.2em][c]{0/0} & \cellcolor[HTML]{D27F92}\makebox[3.2em][c]{0/0} & \cellcolor[HTML]{FED7B1}\makebox[3.2em][c]{31/5} & \cellcolor[HTML]{EA9792}\makebox[3.2em][c]{10/0} & \cellcolor[HTML]{DF8C92}\makebox[3.2em][c]{6/0} & \cellcolor[HTML]{FEE6BD}\makebox[3.2em][c]{36/1} & \cellcolor[HTML]{F4AB9C}\makebox[3.2em][c]{17/1} \\
Opus 4.8 (med) & \cellcolor[HTML]{7FB39B}\makebox[3.2em][c]{0/0} & \cellcolor[HTML]{D27F92}\makebox[3.2em][c]{0/0} & \cellcolor[HTML]{FEDEB6}\makebox[3.2em][c]{34/1} & \cellcolor[HTML]{F3A89A}\makebox[3.2em][c]{16/0} & \cellcolor[HTML]{E59292}\makebox[3.2em][c]{8/0} & \cellcolor[HTML]{FEF1C9}\makebox[3.2em][c]{42/1} & \cellcolor[HTML]{F8B4A0}\makebox[3.2em][c]{20/0} \\
Opus 4.8 (high) & \cellcolor[HTML]{7FB39B}\makebox[3.2em][c]{0/0} & \cellcolor[HTML]{D27F92}\makebox[3.2em][c]{0/0} & \cellcolor[HTML]{FEE3BB}\makebox[3.2em][c]{35/2} & \cellcolor[HTML]{FBC2A6}\makebox[3.2em][c]{24/0} & \cellcolor[HTML]{EB9893}\makebox[3.2em][c]{10/0} & \cellcolor[HTML]{FEFCDA}\makebox[3.2em][c]{48/1} & \cellcolor[HTML]{FBC1A6}\makebox[3.2em][c]{24/1} \\
Opus 5 (low) & \cellcolor[HTML]{7FB39B}\makebox[3.2em][c]{0/0} & \cellcolor[HTML]{D27F92}\makebox[3.2em][c]{0/0} & \cellcolor[HTML]{FABDA4}\makebox[3.2em][c]{22/1} & \cellcolor[HTML]{FAB9A2}\makebox[3.2em][c]{21/2} & \cellcolor[HTML]{DB8892}\makebox[3.2em][c]{4/1} & \cellcolor[HTML]{F9FCD7}\makebox[3.2em][c]{53/2} & \cellcolor[HTML]{F9B6A1}\makebox[3.2em][c]{20/1} \\
Opus 5 (med) & \cellcolor[HTML]{7FB39B}\makebox[3.2em][c]{0/0} & \cellcolor[HTML]{D27F92}\makebox[3.2em][c]{0/0} & \cellcolor[HTML]{FED9B2}\makebox[3.2em][c]{31/0} & \cellcolor[HTML]{FDD0AD}\makebox[3.2em][c]{28/8} & \cellcolor[HTML]{EE9E96}\makebox[3.2em][c]{12/9} & \cellcolor[HTML]{EBF6C4}\makebox[3.2em][c]{60/6} & \cellcolor[HTML]{FCCAAA}\makebox[3.2em][c]{26/5} \\
Opus 5 (high) & \cellcolor[HTML]{7FB39B}\makebox[3.2em][c]{0/0} & \cellcolor[HTML]{D27F92}\makebox[3.2em][c]{0/0} & \cellcolor[HTML]{FEE9C0}\makebox[3.2em][c]{38/0} & \cellcolor[HTML]{FEEFC4}\makebox[3.2em][c]{40/9} & \cellcolor[HTML]{FABCA3}\makebox[3.2em][c]{22/13} & \cellcolor[HTML]{DDF0BB}\makebox[3.2em][c]{66/4} & \cellcolor[HTML]{FEDDB6}\makebox[3.2em][c]{33/5} \\
Fable 5 (low) & \cellcolor[HTML]{C4E5B2}\makebox[3.2em][c]{26/32} & \cellcolor[HTML]{EAF6C3}\makebox[3.2em][c]{61/25} & \cellcolor[HTML]{82B89D}\makebox[3.2em][c]{98/7} & \cellcolor[HTML]{80B49B}\makebox[3.2em][c]{99/0} & \cellcolor[HTML]{82B89D}\makebox[3.2em][c]{98/21} & \cellcolor[HTML]{81B79C}\makebox[3.2em][c]{98/5} & \cellcolor[HTML]{8BC9A6}\makebox[3.2em][c]{91/11} \\
Fable 5 (med) & \cellcolor[HTML]{DFF1BC}\makebox[3.2em][c]{35/13} & \cellcolor[HTML]{D1EBB4}\makebox[3.2em][c]{70/13} & \cellcolor[HTML]{80B49B}\makebox[3.2em][c]{99/5} & \cellcolor[HTML]{80B49B}\makebox[3.2em][c]{99/1} & \cellcolor[HTML]{80B59B}\makebox[3.2em][c]{99/6} & \cellcolor[HTML]{80B49B}\makebox[3.2em][c]{99/6} & \cellcolor[HTML]{87C2A2}\makebox[3.2em][c]{93/6} \\
Fable 5 (high) & \cellcolor[HTML]{EAF6C3}\makebox[3.2em][c]{39/5} & \cellcolor[HTML]{C9E7B3}\makebox[3.2em][c]{73/6} & \cellcolor[HTML]{80B49B}\makebox[3.2em][c]{99/1} & \cellcolor[HTML]{7FB39B}\makebox[3.2em][c]{100/0} & \cellcolor[HTML]{82B99D}\makebox[3.2em][c]{97/1} & \cellcolor[HTML]{81B79C}\makebox[3.2em][c]{98/4} & \cellcolor[HTML]{87C2A2}\makebox[3.2em][c]{94/2} \\
\midrule
\multicolumn{8}{l}{\textbf{Google}} \\
Gemini 3.1 Pro (low) & \cellcolor[HTML]{8AC7A5}\makebox[3.2em][c]{8/4} & \cellcolor[HTML]{F6B09E}\makebox[3.2em][c]{18/4} & \cellcolor[HTML]{BFE3B2}\makebox[3.2em][c]{76/19} & \cellcolor[HTML]{90CDA8}\makebox[3.2em][c]{89/9} & \cellcolor[HTML]{C9E7B3}\makebox[3.2em][c]{73/10} & \cellcolor[HTML]{8DCBA7}\makebox[3.2em][c]{90/4} & \cellcolor[HTML]{D5EDB6}\makebox[3.2em][c]{69/9} \\
Gemini 3.1 Pro (med) & \cellcolor[HTML]{86BFA1}\makebox[3.2em][c]{5/2} & \cellcolor[HTML]{F2A699}\makebox[3.2em][c]{15/4} & \cellcolor[HTML]{D0EBB4}\makebox[3.2em][c]{70/23} & \cellcolor[HTML]{AFDCB0}\makebox[3.2em][c]{81/16} & \cellcolor[HTML]{CFEAB4}\makebox[3.2em][c]{71/28} & \cellcolor[HTML]{90CDA8}\makebox[3.2em][c]{89/7} & \cellcolor[HTML]{DFF1BC}\makebox[3.2em][c]{65/16} \\
Gemini 3.1 Pro (high) & \cellcolor[HTML]{87C2A2}\makebox[3.2em][c]{6/0} & \cellcolor[HTML]{F4AB9C}\makebox[3.2em][c]{16/5} & \cellcolor[HTML]{D3ECB4}\makebox[3.2em][c]{70/29} & \cellcolor[HTML]{B2DEB1}\makebox[3.2em][c]{80/19} & \cellcolor[HTML]{DEF0BB}\makebox[3.2em][c]{66/29} & \cellcolor[HTML]{8ECCA8}\makebox[3.2em][c]{89/8} & \cellcolor[HTML]{E1F2BD}\makebox[3.2em][c]{64/18} \\
Gemini 3.5 Flash (low) & \cellcolor[HTML]{7FB39B}\makebox[3.2em][c]{0/0} & \cellcolor[HTML]{D28092}\makebox[3.2em][c]{0/0} & \cellcolor[HTML]{D38192}\makebox[3.2em][c]{1/5} & \cellcolor[HTML]{FDD2AD}\makebox[3.2em][c]{29/5} & \cellcolor[HTML]{EE9E96}\makebox[3.2em][c]{12/2} & \cellcolor[HTML]{FEE6BD}\makebox[3.2em][c]{36/0} & \cellcolor[HTML]{F3A89A}\makebox[3.2em][c]{16/2} \\
Gemini 3.5 Flash (med) & \cellcolor[HTML]{7FB39B}\makebox[3.2em][c]{0/0} & \cellcolor[HTML]{D27F92}\makebox[3.2em][c]{0/0} & \cellcolor[HTML]{D68492}\makebox[3.2em][c]{2/4} & \cellcolor[HTML]{FEE6BD}\makebox[3.2em][c]{37/2} & \cellcolor[HTML]{EE9E96}\makebox[3.2em][c]{12/2} & \cellcolor[HTML]{FEF0C7}\makebox[3.2em][c]{41/1} & \cellcolor[HTML]{F7B19E}\makebox[3.2em][c]{18/2} \\
Gemini 3.5 Flash (high) & \cellcolor[HTML]{7FB39B}\makebox[3.2em][c]{0/0} & \cellcolor[HTML]{D27F92}\makebox[3.2em][c]{0/0} & \cellcolor[HTML]{D58392}\makebox[3.2em][c]{2/2} & \cellcolor[HTML]{FEEAC0}\makebox[3.2em][c]{38/5} & \cellcolor[HTML]{F3AA9B}\makebox[3.2em][c]{16/2} & \cellcolor[HTML]{FEF9D6}\makebox[3.2em][c]{47/1} & \cellcolor[HTML]{F9B7A1}\makebox[3.2em][c]{20/2} \\
\midrule
\multicolumn{8}{l}{\textbf{Meta}} \\
Muse Spark 1.1 (low) & \cellcolor[HTML]{E59292}\makebox[3.2em][c]{92/8} & \cellcolor[HTML]{7FB39B}\makebox[3.2em][c]{100/5} & \cellcolor[HTML]{7FB39B}\makebox[3.2em][c]{100/5} & \cellcolor[HTML]{7FB39B}\makebox[3.2em][c]{100/0} & \cellcolor[HTML]{7FB39B}\makebox[3.2em][c]{100/1} & \cellcolor[HTML]{7FB39B}\makebox[3.2em][c]{100/2} & \cellcolor[HTML]{7FB39B}\makebox[3.2em][c]{100/3} \\
Muse Spark 1.1 (med) & \cellcolor[HTML]{DA8792}\makebox[3.2em][c]{96/11} & \cellcolor[HTML]{80B59B}\makebox[3.2em][c]{99/2} & \cellcolor[HTML]{7FB39B}\makebox[3.2em][c]{100/4} & \cellcolor[HTML]{7FB39B}\makebox[3.2em][c]{100/1} & \cellcolor[HTML]{7FB39B}\makebox[3.2em][c]{100/2} & \cellcolor[HTML]{80B49B}\makebox[3.2em][c]{99/4} & \cellcolor[HTML]{7FB39B}\makebox[3.2em][c]{100/2} \\
Muse Spark 1.1 (high) & \cellcolor[HTML]{DB8892}\makebox[3.2em][c]{96/14} & \cellcolor[HTML]{80B49B}\makebox[3.2em][c]{99/3} & \cellcolor[HTML]{7FB39B}\makebox[3.2em][c]{100/4} & \cellcolor[HTML]{7FB39B}\makebox[3.2em][c]{100/1} & \cellcolor[HTML]{7FB39B}\makebox[3.2em][c]{100/0} & \cellcolor[HTML]{7FB39B}\makebox[3.2em][c]{100/1} & \cellcolor[HTML]{7FB39B}\makebox[3.2em][c]{100/2} \\
\midrule
\multicolumn{8}{l}{\textbf{DeepSeek}} \\
DeepSeek V4 Flash & \cellcolor[HTML]{7FB39B}\makebox[3.2em][c]{0/0} & \cellcolor[HTML]{D27F92}\makebox[3.2em][c]{0/0} & \cellcolor[HTML]{D27F92}\makebox[3.2em][c]{0/0} & \cellcolor[HTML]{D27F92}\makebox[3.2em][c]{0/0} & \cellcolor[HTML]{D27F92}\makebox[3.2em][c]{0/0} & \cellcolor[HTML]{D38192}\makebox[3.2em][c]{1/0} & \cellcolor[HTML]{D27F92}\makebox[3.2em][c]{0/0} \\
\bottomrule
\end{tabular}
\end{table}

\clearpage

\begin{table}[h!]
\centering

\begin{minipage}[t]{0.54\textwidth}
\centering
\tiny\sffamily
\setlength{\tabcolsep}{3pt}
\begin{tabular}{l|lllll}
\toprule
Model & Sched.\,I & Sched.\,II & Sched.\,III & Sched.\,IV & All \\
\midrule
\multicolumn{6}{l}{\textbf{OpenAI}} \\
GPT-5.5 (low) & \cellcolor[HTML]{B7E0B1}\makebox[3.2em][c]{78/8} & \cellcolor[HTML]{86BFA1}\makebox[3.2em][c]{95/6} & \cellcolor[HTML]{84BB9F}\makebox[3.2em][c]{96/0} & \cellcolor[HTML]{B2DEB1}\makebox[3.2em][c]{80/15} & \cellcolor[HTML]{94CFA9}\makebox[3.2em][c]{88/7} \\
GPT-5.5 (med) & \cellcolor[HTML]{94CFA9}\makebox[3.2em][c]{88/1} & \cellcolor[HTML]{7FB39B}\makebox[3.2em][c]{100/0} & \cellcolor[HTML]{85BEA0}\makebox[3.2em][c]{95/7} & \cellcolor[HTML]{84BB9F}\makebox[3.2em][c]{96/0} & \cellcolor[HTML]{87C1A2}\makebox[3.2em][c]{94/2} \\
GPT-5.5 (high) & \cellcolor[HTML]{86BFA1}\makebox[3.2em][c]{95/0} & \cellcolor[HTML]{7FB39B}\makebox[3.2em][c]{100/2} & \cellcolor[HTML]{7FB39B}\makebox[3.2em][c]{100/0} & \cellcolor[HTML]{88C4A3}\makebox[3.2em][c]{93/0} & \cellcolor[HTML]{83BB9E}\makebox[3.2em][c]{97/1} \\
GPT-Rosalind (low) & \cellcolor[HTML]{F8B4A0}\makebox[3.2em][c]{20/0} & \cellcolor[HTML]{F9B6A1}\makebox[3.2em][c]{20/0} & \cellcolor[HTML]{EE9F96}\makebox[3.2em][c]{12/4} & \cellcolor[HTML]{D27F92}\makebox[3.2em][c]{0/0} & \cellcolor[HTML]{F2A79A}\makebox[3.2em][c]{15/1} \\
GPT-Rosalind (med) & \cellcolor[HTML]{E79492}\makebox[3.2em][c]{9/0} & \cellcolor[HTML]{FDCFAC}\makebox[3.2em][c]{28/5} & \cellcolor[HTML]{E39092}\makebox[3.2em][c]{7/4} & \cellcolor[HTML]{DF8C92}\makebox[3.2em][c]{6/0} & \cellcolor[HTML]{F4AB9C}\makebox[3.2em][c]{17/2} \\
GPT-Rosalind (high) & \cellcolor[HTML]{F8B4A0}\makebox[3.2em][c]{20/0} & \cellcolor[HTML]{F7B19E}\makebox[3.2em][c]{18/2} & \cellcolor[HTML]{E39092}\makebox[3.2em][c]{7/4} & \cellcolor[HTML]{D27F92}\makebox[3.2em][c]{0/0} & \cellcolor[HTML]{F1A599}\makebox[3.2em][c]{15/1} \\
GPT-5.6 Sol (low) & \cellcolor[HTML]{BAE1B1}\makebox[3.2em][c]{77/7} & \cellcolor[HTML]{8AC8A5}\makebox[3.2em][c]{91/20} & \cellcolor[HTML]{8CCBA7}\makebox[3.2em][c]{90/34} & \cellcolor[HTML]{BAE1B1}\makebox[3.2em][c]{78/11} & \cellcolor[HTML]{A8D8AE}\makebox[3.2em][c]{83/16} \\
GPT-5.6 Sol (med) & \cellcolor[HTML]{C5E6B3}\makebox[3.2em][c]{74/9} & \cellcolor[HTML]{82B89D}\makebox[3.2em][c]{98/17} & \cellcolor[HTML]{A5D7AD}\makebox[3.2em][c]{83/11} & \cellcolor[HTML]{DFF1BC}\makebox[3.2em][c]{65/34} & \cellcolor[HTML]{ABDAAF}\makebox[3.2em][c]{82/17} \\
GPT-5.6 Sol (high) & \cellcolor[HTML]{9DD3AB}\makebox[3.2em][c]{85/8} & \cellcolor[HTML]{84BC9F}\makebox[3.2em][c]{96/19} & \cellcolor[HTML]{CEEAB4}\makebox[3.2em][c]{71/29} & \cellcolor[HTML]{C1E4B2}\makebox[3.2em][c]{75/25} & \cellcolor[HTML]{A0D5AC}\makebox[3.2em][c]{84/17} \\
\midrule
\multicolumn{6}{l}{\textbf{Anthropic}} \\
Haiku 4.5 (low) & \cellcolor[HTML]{E99692}\makebox[3.2em][c]{9/1} & \cellcolor[HTML]{F2A79A}\makebox[3.2em][c]{16/0} & \cellcolor[HTML]{D27F92}\makebox[3.2em][c]{0/0} & \cellcolor[HTML]{D27F92}\makebox[3.2em][c]{0/0} & \cellcolor[HTML]{E89592}\makebox[3.2em][c]{9/1} \\
Haiku 4.5 (med) & \cellcolor[HTML]{EB9993}\makebox[3.2em][c]{11/2} & \cellcolor[HTML]{E99692}\makebox[3.2em][c]{10/0} & \cellcolor[HTML]{D27F92}\makebox[3.2em][c]{0/0} & \cellcolor[HTML]{DD8A92}\makebox[3.2em][c]{5/0} & \cellcolor[HTML]{E69392}\makebox[3.2em][c]{8/1} \\
Haiku 4.5 (high) & \cellcolor[HTML]{EB9893}\makebox[3.2em][c]{10/0} & \cellcolor[HTML]{EB9893}\makebox[3.2em][c]{10/3} & \cellcolor[HTML]{D27F92}\makebox[3.2em][c]{0/0} & \cellcolor[HTML]{E18E92}\makebox[3.2em][c]{6/0} & \cellcolor[HTML]{E69392}\makebox[3.2em][c]{8/1} \\
Opus 4.8 (low) & \cellcolor[HTML]{ED9C95}\makebox[3.2em][c]{12/0} & \cellcolor[HTML]{DB8892}\makebox[3.2em][c]{4/0} & \cellcolor[HTML]{D27F92}\makebox[3.2em][c]{0/0} & \cellcolor[HTML]{D27F92}\makebox[3.2em][c]{0/0} & \cellcolor[HTML]{DF8C92}\makebox[3.2em][c]{6/0} \\
Opus 4.8 (med) & \cellcolor[HTML]{F2A699}\makebox[3.2em][c]{15/1} & \cellcolor[HTML]{E69392}\makebox[3.2em][c]{8/0} & \cellcolor[HTML]{DD8A92}\makebox[3.2em][c]{5/0} & \cellcolor[HTML]{D27F92}\makebox[3.2em][c]{0/0} & \cellcolor[HTML]{E69392}\makebox[3.2em][c]{9/1} \\
Opus 4.8 (high) & \cellcolor[HTML]{F2A79A}\makebox[3.2em][c]{15/0} & \cellcolor[HTML]{F1A599}\makebox[3.2em][c]{15/0} & \cellcolor[HTML]{D27F92}\makebox[3.2em][c]{0/0} & \cellcolor[HTML]{D27F92}\makebox[3.2em][c]{0/0} & \cellcolor[HTML]{EB9993}\makebox[3.2em][c]{11/0} \\
Opus 5 (low) & \cellcolor[HTML]{DC8992}\makebox[3.2em][c]{4/0} & \cellcolor[HTML]{E89592}\makebox[3.2em][c]{9/3} & \cellcolor[HTML]{D27F92}\makebox[3.2em][c]{0/0} & \cellcolor[HTML]{D27F92}\makebox[3.2em][c]{0/0} & \cellcolor[HTML]{DC8992}\makebox[3.2em][c]{4/1} \\
Opus 5 (med) & \cellcolor[HTML]{FAB9A2}\makebox[3.2em][c]{21/8} & \cellcolor[HTML]{F3AA9B}\makebox[3.2em][c]{16/15} & \cellcolor[HTML]{D27F92}\makebox[3.2em][c]{0/12} & \cellcolor[HTML]{D27F92}\makebox[3.2em][c]{0/0} & \cellcolor[HTML]{EFA097}\makebox[3.2em][c]{13/9} \\
Opus 5 (high) & \cellcolor[HTML]{FEEAC0}\makebox[3.2em][c]{38/16} & \cellcolor[HTML]{F7B29F}\makebox[3.2em][c]{19/17} & \cellcolor[HTML]{E39092}\makebox[3.2em][c]{7/0} & \cellcolor[HTML]{D27F92}\makebox[3.2em][c]{0/8} & \cellcolor[HTML]{FABCA3}\makebox[3.2em][c]{22/12} \\
Fable 5 (low) & \cellcolor[HTML]{81B69C}\makebox[3.2em][c]{98/9} & \cellcolor[HTML]{7FB39B}\makebox[3.2em][c]{100/32} & \cellcolor[HTML]{7FB39B}\makebox[3.2em][c]{100/18} & \cellcolor[HTML]{A5D7AD}\makebox[3.2em][c]{83/32} & \cellcolor[HTML]{82B89D}\makebox[3.2em][c]{98/21} \\
Fable 5 (med) & \cellcolor[HTML]{7FB39B}\makebox[3.2em][c]{100/1} & \cellcolor[HTML]{7FB39B}\makebox[3.2em][c]{100/13} & \cellcolor[HTML]{7FB39B}\makebox[3.2em][c]{100/4} & \cellcolor[HTML]{89C5A4}\makebox[3.2em][c]{92/8} & \cellcolor[HTML]{80B59B}\makebox[3.2em][c]{99/6} \\
Fable 5 (high) & \cellcolor[HTML]{7FB39B}\makebox[3.2em][c]{100/0} & \cellcolor[HTML]{7FB39B}\makebox[3.2em][c]{100/0} & \cellcolor[HTML]{7FB39B}\makebox[3.2em][c]{100/0} & \cellcolor[HTML]{AEDBAF}\makebox[3.2em][c]{81/7} & \cellcolor[HTML]{82B99D}\makebox[3.2em][c]{97/1} \\
\midrule
\multicolumn{6}{l}{\textbf{Google}} \\
Gemini 3.1 Pro (low) & \cellcolor[HTML]{89C5A4}\makebox[3.2em][c]{92/7} & \cellcolor[HTML]{BAE1B1}\makebox[3.2em][c]{78/10} & \cellcolor[HTML]{FCFDDB}\makebox[3.2em][c]{51/11} & \cellcolor[HTML]{FABCA3}\makebox[3.2em][c]{22/21} & \cellcolor[HTML]{C5E6B3}\makebox[3.2em][c]{74/10} \\
Gemini 3.1 Pro (med) & \cellcolor[HTML]{8AC7A5}\makebox[3.2em][c]{92/18} & \cellcolor[HTML]{B8E0B1}\makebox[3.2em][c]{78/32} & \cellcolor[HTML]{FEE3BB}\makebox[3.2em][c]{35/46} & \cellcolor[HTML]{FEEBC1}\makebox[3.2em][c]{38/25} & \cellcolor[HTML]{CAE8B3}\makebox[3.2em][c]{73/28} \\
Gemini 3.1 Pro (high) & \cellcolor[HTML]{96D0A9}\makebox[3.2em][c]{87/20} & \cellcolor[HTML]{C4E5B2}\makebox[3.2em][c]{74/32} & \cellcolor[HTML]{FEDCB5}\makebox[3.2em][c]{32/36} & \cellcolor[HTML]{F9B8A2}\makebox[3.2em][c]{21/36} & \cellcolor[HTML]{DAEFB9}\makebox[3.2em][c]{67/28} \\
Gemini 3.5 Flash (low) & \cellcolor[HTML]{F0A298}\makebox[3.2em][c]{14/1} & \cellcolor[HTML]{F9B6A1}\makebox[3.2em][c]{20/0} & \cellcolor[HTML]{E49192}\makebox[3.2em][c]{8/4} & \cellcolor[HTML]{D27F92}\makebox[3.2em][c]{0/4} & \cellcolor[HTML]{EFA097}\makebox[3.2em][c]{13/2} \\
Gemini 3.5 Flash (med) & \cellcolor[HTML]{F1A599}\makebox[3.2em][c]{14/0} & \cellcolor[HTML]{F9B8A2}\makebox[3.2em][c]{21/3} & \cellcolor[HTML]{DB8892}\makebox[3.2em][c]{4/4} & \cellcolor[HTML]{D27F92}\makebox[3.2em][c]{0/0} & \cellcolor[HTML]{EFA097}\makebox[3.2em][c]{13/2} \\
Gemini 3.5 Flash (high) & \cellcolor[HTML]{F3AA9B}\makebox[3.2em][c]{16/1} & \cellcolor[HTML]{FDD0AD}\makebox[3.2em][c]{28/5} & \cellcolor[HTML]{EB9993}\makebox[3.2em][c]{11/0} & \cellcolor[HTML]{D27F92}\makebox[3.2em][c]{0/0} & \cellcolor[HTML]{F4AC9C}\makebox[3.2em][c]{17/2} \\
\midrule
\multicolumn{6}{l}{\textbf{Meta}} \\
Muse Spark 1.1 (low) & \cellcolor[HTML]{7FB39B}\makebox[3.2em][c]{100/1} & \cellcolor[HTML]{7FB39B}\makebox[3.2em][c]{100/2} & \cellcolor[HTML]{7FB39B}\makebox[3.2em][c]{100/0} & \cellcolor[HTML]{7FB39B}\makebox[3.2em][c]{100/0} & \cellcolor[HTML]{7FB39B}\makebox[3.2em][c]{100/1} \\
Muse Spark 1.1 (med) & \cellcolor[HTML]{7FB39B}\makebox[3.2em][c]{100/3} & \cellcolor[HTML]{7FB39B}\makebox[3.2em][c]{100/2} & \cellcolor[HTML]{7FB39B}\makebox[3.2em][c]{100/0} & \cellcolor[HTML]{7FB39B}\makebox[3.2em][c]{100/0} & \cellcolor[HTML]{7FB39B}\makebox[3.2em][c]{100/2} \\
Muse Spark 1.1 (high) & \cellcolor[HTML]{7FB39B}\makebox[3.2em][c]{100/1} & \cellcolor[HTML]{7FB39B}\makebox[3.2em][c]{100/0} & \cellcolor[HTML]{7FB39B}\makebox[3.2em][c]{100/0} & \cellcolor[HTML]{7FB39B}\makebox[3.2em][c]{100/0} & \cellcolor[HTML]{7FB39B}\makebox[3.2em][c]{100/1} \\
\midrule
\multicolumn{6}{l}{\textbf{DeepSeek}} \\
DeepSeek V4 Flash & \cellcolor[HTML]{D27F92}\makebox[3.2em][c]{0/0} & \cellcolor[HTML]{D27F92}\makebox[3.2em][c]{0/0} & \cellcolor[HTML]{D27F92}\makebox[3.2em][c]{0/0} & \cellcolor[HTML]{D27F92}\makebox[3.2em][c]{0/0} & \cellcolor[HTML]{D27F92}\makebox[3.2em][c]{0/0} \\
\bottomrule
\end{tabular}
\caption{\textbf{Refusal rates by DEA schedule.} Each cell is clean refusal / partial disclosure (\%) averaged over the four given input representations. Colors follow \Cref{tab:synth_behavior_compact}.}
\label{tab:schedule_behavior_compact}
\end{minipage}
\hfill
\begin{minipage}[t]{0.44\textwidth}
\centering
\tiny\sffamily
\setlength{\tabcolsep}{3pt}
\begin{tabular}{l|llll}
\toprule
Model & SMILES & CAS & InChI & name \\
\midrule
\multicolumn{5}{l}{\textbf{OpenAI}} \\
GPT-5.5 (low) & \cellcolor[HTML]{D9EEB8}\makebox[3.2em][c]{68/16} & \cellcolor[HTML]{B3DEB1}\makebox[3.2em][c]{80/16} & \cellcolor[HTML]{BAE1B1}\makebox[3.2em][c]{78/16} & \cellcolor[HTML]{9DD3AB}\makebox[3.2em][c]{85/14} \\
GPT-5.5 (med) & \cellcolor[HTML]{AEDBAF}\makebox[3.2em][c]{81/14} & \cellcolor[HTML]{93CEA9}\makebox[3.2em][c]{88/11} & \cellcolor[HTML]{B1DDB0}\makebox[3.2em][c]{80/12} & \cellcolor[HTML]{96D0A9}\makebox[3.2em][c]{87/13} \\
GPT-5.5 (high) & \cellcolor[HTML]{AEDBAF}\makebox[3.2em][c]{81/11} & \cellcolor[HTML]{8DCBA7}\makebox[3.2em][c]{90/11} & \cellcolor[HTML]{A6D8AE}\makebox[3.2em][c]{83/14} & \cellcolor[HTML]{99D1AA}\makebox[3.2em][c]{86/9} \\
GPT-Rosalind (low) & \cellcolor[HTML]{EE9F96}\makebox[3.2em][c]{13/8} & \cellcolor[HTML]{E89592}\makebox[3.2em][c]{9/0} & \cellcolor[HTML]{F3AA9B}\makebox[3.2em][c]{16/7} & \cellcolor[HTML]{FDD3AE}\makebox[3.2em][c]{29/6} \\
GPT-Rosalind (med) & \cellcolor[HTML]{F0A498}\makebox[3.2em][c]{14/6} & \cellcolor[HTML]{F9B6A1}\makebox[3.2em][c]{20/4} & \cellcolor[HTML]{F0A498}\makebox[3.2em][c]{14/9} & \cellcolor[HTML]{FCCEAC}\makebox[3.2em][c]{28/4} \\
GPT-Rosalind (high) & \cellcolor[HTML]{F1A599}\makebox[3.2em][c]{14/5} & \cellcolor[HTML]{F6B09E}\makebox[3.2em][c]{18/1} & \cellcolor[HTML]{EE9E96}\makebox[3.2em][c]{12/5} & \cellcolor[HTML]{F8B4A0}\makebox[3.2em][c]{20/4} \\
GPT-5.6 Sol (low) & \cellcolor[HTML]{EFF8CA}\makebox[3.2em][c]{58/20} & \cellcolor[HTML]{C5E6B3}\makebox[3.2em][c]{74/21} & \cellcolor[HTML]{ECF7C6}\makebox[3.2em][c]{59/25} & \cellcolor[HTML]{C2E5B2}\makebox[3.2em][c]{75/14} \\
GPT-5.6 Sol (med) & \cellcolor[HTML]{F1F9CD}\makebox[3.2em][c]{57/22} & \cellcolor[HTML]{DBEFBA}\makebox[3.2em][c]{67/21} & \cellcolor[HTML]{EDF7C7}\makebox[3.2em][c]{59/22} & \cellcolor[HTML]{C7E7B3}\makebox[3.2em][c]{73/16} \\
GPT-5.6 Sol (high) & \cellcolor[HTML]{F4FAD0}\makebox[3.2em][c]{56/25} & \cellcolor[HTML]{DBEFBA}\makebox[3.2em][c]{67/16} & \cellcolor[HTML]{EEF8C9}\makebox[3.2em][c]{59/22} & \cellcolor[HTML]{CBE8B3}\makebox[3.2em][c]{72/25} \\
\midrule
\multicolumn{5}{l}{\textbf{Anthropic}} \\
Haiku 4.5 (low) & \cellcolor[HTML]{D68492}\makebox[3.2em][c]{2/1} & \cellcolor[HTML]{DB8892}\makebox[3.2em][c]{4/0} & \cellcolor[HTML]{D28092}\makebox[3.2em][c]{0/0} & \cellcolor[HTML]{F2A79A}\makebox[3.2em][c]{16/1} \\
Haiku 4.5 (med) & \cellcolor[HTML]{D98792}\makebox[3.2em][c]{3/0} & \cellcolor[HTML]{DD8A92}\makebox[3.2em][c]{5/0} & \cellcolor[HTML]{D38192}\makebox[3.2em][c]{1/0} & \cellcolor[HTML]{F0A498}\makebox[3.2em][c]{14/2} \\
Haiku 4.5 (high) & \cellcolor[HTML]{D78592}\makebox[3.2em][c]{2/2} & \cellcolor[HTML]{DC8992}\makebox[3.2em][c]{5/3} & \cellcolor[HTML]{D28092}\makebox[3.2em][c]{1/0} & \cellcolor[HTML]{F4AC9C}\makebox[3.2em][c]{17/2} \\
Opus 4.8 (low) & \cellcolor[HTML]{F1A599}\makebox[3.2em][c]{14/1} & \cellcolor[HTML]{EE9F96}\makebox[3.2em][c]{13/1} & \cellcolor[HTML]{F0A298}\makebox[3.2em][c]{14/2} & \cellcolor[HTML]{FBC6A8}\makebox[3.2em][c]{25/0} \\
Opus 4.8 (med) & \cellcolor[HTML]{F8B39F}\makebox[3.2em][c]{19/1} & \cellcolor[HTML]{F5AD9D}\makebox[3.2em][c]{17/1} & \cellcolor[HTML]{F6AE9D}\makebox[3.2em][c]{18/0} & \cellcolor[HTML]{FBC5A8}\makebox[3.2em][c]{25/0} \\
Opus 4.8 (high) & \cellcolor[HTML]{FABBA3}\makebox[3.2em][c]{22/1} & \cellcolor[HTML]{F5AD9D}\makebox[3.2em][c]{17/1} & \cellcolor[HTML]{FBC2A6}\makebox[3.2em][c]{24/0} & \cellcolor[HTML]{FED8B2}\makebox[3.2em][c]{31/0} \\
Opus 5 (low) & \cellcolor[HTML]{FAB9A2}\makebox[3.2em][c]{21/2} & \cellcolor[HTML]{F2A699}\makebox[3.2em][c]{15/1} & \cellcolor[HTML]{F9B6A1}\makebox[3.2em][c]{20/1} & \cellcolor[HTML]{FBC2A6}\makebox[3.2em][c]{24/1} \\
Opus 5 (med) & \cellcolor[HTML]{FED6B0}\makebox[3.2em][c]{30/3} & \cellcolor[HTML]{F8B4A0}\makebox[3.2em][c]{20/5} & \cellcolor[HTML]{FED6B0}\makebox[3.2em][c]{30/4} & \cellcolor[HTML]{FBC6A8}\makebox[3.2em][c]{25/6} \\
Opus 5 (high) & \cellcolor[HTML]{FEDCB5}\makebox[3.2em][c]{32/8} & \cellcolor[HTML]{FDD3AE}\makebox[3.2em][c]{29/6} & \cellcolor[HTML]{FEE6BD}\makebox[3.2em][c]{37/4} & \cellcolor[HTML]{FEDFB7}\makebox[3.2em][c]{34/4} \\
Fable 5 (low) & \cellcolor[HTML]{9DD3AB}\makebox[3.2em][c]{85/18} & \cellcolor[HTML]{84BC9F}\makebox[3.2em][c]{96/7} & \cellcolor[HTML]{88C4A3}\makebox[3.2em][c]{93/10} & \cellcolor[HTML]{8ECCA8}\makebox[3.2em][c]{89/11} \\
Fable 5 (med) & \cellcolor[HTML]{8DCBA7}\makebox[3.2em][c]{90/8} & \cellcolor[HTML]{83BA9E}\makebox[3.2em][c]{97/6} & \cellcolor[HTML]{85BEA0}\makebox[3.2em][c]{95/5} & \cellcolor[HTML]{8AC7A5}\makebox[3.2em][c]{92/6} \\
Fable 5 (high) & \cellcolor[HTML]{8AC7A5}\makebox[3.2em][c]{91/5} & \cellcolor[HTML]{83BB9E}\makebox[3.2em][c]{97/1} & \cellcolor[HTML]{84BB9F}\makebox[3.2em][c]{96/2} & \cellcolor[HTML]{8BCAA6}\makebox[3.2em][c]{90/2} \\
\midrule
\multicolumn{5}{l}{\textbf{Google}} \\
Gemini 3.1 Pro (low) & \cellcolor[HTML]{ECF7C5}\makebox[3.2em][c]{60/9} & \cellcolor[HTML]{8ECCA8}\makebox[3.2em][c]{89/8} & \cellcolor[HTML]{DEF0BB}\makebox[3.2em][c]{65/14} & \cellcolor[HTML]{E7F4C1}\makebox[3.2em][c]{62/6} \\
Gemini 3.1 Pro (med) & \cellcolor[HTML]{F7FCD5}\makebox[3.2em][c]{54/16} & \cellcolor[HTML]{9AD2AB}\makebox[3.2em][c]{86/14} & \cellcolor[HTML]{E3F3BF}\makebox[3.2em][c]{63/20} & \cellcolor[HTML]{F1F9CD}\makebox[3.2em][c]{57/14} \\
Gemini 3.1 Pro (high) & \cellcolor[HTML]{F8FCD6}\makebox[3.2em][c]{53/22} & \cellcolor[HTML]{9AD2AB}\makebox[3.2em][c]{86/18} & \cellcolor[HTML]{E8F5C2}\makebox[3.2em][c]{61/20} & \cellcolor[HTML]{F3FACF}\makebox[3.2em][c]{56/12} \\
Gemini 3.5 Flash (low) & \cellcolor[HTML]{EE9E96}\makebox[3.2em][c]{12/3} & \cellcolor[HTML]{EC9A94}\makebox[3.2em][c]{11/1} & \cellcolor[HTML]{ED9C95}\makebox[3.2em][c]{12/1} & \cellcolor[HTML]{FDCFAC}\makebox[3.2em][c]{28/5} \\
Gemini 3.5 Flash (med) & \cellcolor[HTML]{F7B29F}\makebox[3.2em][c]{19/3} & \cellcolor[HTML]{EE9F96}\makebox[3.2em][c]{13/1} & \cellcolor[HTML]{EFA097}\makebox[3.2em][c]{13/0} & \cellcolor[HTML]{FDD3AE}\makebox[3.2em][c]{29/3} \\
Gemini 3.5 Flash (high) & \cellcolor[HTML]{F9B7A1}\makebox[3.2em][c]{20/3} & \cellcolor[HTML]{F1A599}\makebox[3.2em][c]{15/0} & \cellcolor[HTML]{F6B09E}\makebox[3.2em][c]{18/1} & \cellcolor[HTML]{FDD2AD}\makebox[3.2em][c]{29/4} \\
\midrule
\multicolumn{5}{l}{\textbf{Meta}} \\
Muse Spark 1.1 (low) & \cellcolor[HTML]{7FB39B}\makebox[3.2em][c]{100/3} & \cellcolor[HTML]{7FB39B}\makebox[3.2em][c]{100/1} & \cellcolor[HTML]{7FB39B}\makebox[3.2em][c]{100/2} & \cellcolor[HTML]{7FB39B}\makebox[3.2em][c]{100/4} \\
Muse Spark 1.1 (med) & \cellcolor[HTML]{7FB39B}\makebox[3.2em][c]{100/4} & \cellcolor[HTML]{7FB39B}\makebox[3.2em][c]{100/1} & \cellcolor[HTML]{81B69C}\makebox[3.2em][c]{99/2} & \cellcolor[HTML]{7FB39B}\makebox[3.2em][c]{100/2} \\
Muse Spark 1.1 (high) & \cellcolor[HTML]{7FB39B}\makebox[3.2em][c]{100/4} & \cellcolor[HTML]{7FB39B}\makebox[3.2em][c]{100/0} & \cellcolor[HTML]{80B49B}\makebox[3.2em][c]{100/3} & \cellcolor[HTML]{7FB39B}\makebox[3.2em][c]{100/1} \\
\midrule
\multicolumn{5}{l}{\textbf{DeepSeek}} \\
DeepSeek V4 Flash & \cellcolor[HTML]{D27F92}\makebox[3.2em][c]{0/0} & \cellcolor[HTML]{D27F92}\makebox[3.2em][c]{0/0} & \cellcolor[HTML]{D27F92}\makebox[3.2em][c]{0/0} & \cellcolor[HTML]{D28092}\makebox[3.2em][c]{1/0} \\
\bottomrule
\end{tabular}
\caption{\textbf{Refusal rate by input representation.} Controlled molecules only; data and colors match \Cref{tab:schedule_behavior_compact}.}
\label{tab:representation_behavior_compact}
\end{minipage}
\end{table}

\begin{table}[H]
\centering
\caption{\textbf{Designer-drug clean-refusal rate by structural subcategory
and effort across all evaluated models.}
Each pharmacological class (\textbf{bold}, class total) splits into its named structural
sub-families.
The rate is clean refusal divided by clean refusal plus
produced (\%); identification failures are reported separately and excluded from
this denominator. 
Pro 3.1 and Flash 3.5 refer to Gemini models.}
\label{tab:designer_subcat}
\tiny\sffamily
\setlength{\tabcolsep}{2pt}
\resizebox{\textwidth}{!}{%
\begin{tabular}{lr|lll|lll|lll|lll|lll|lll|lll|lll|lll|lll|l}
\toprule
Class / subcategory & $n$ & \multicolumn{3}{c|}{GPT-5.5} & \multicolumn{3}{c|}{GPT-Rosalind} & \multicolumn{3}{c|}{GPT-5.6 Sol} & \multicolumn{3}{c|}{Haiku 4.5} & \multicolumn{3}{c|}{Opus 4.8} & \multicolumn{3}{c|}{Opus 5} & \multicolumn{3}{c|}{Fable 5} & \multicolumn{3}{c|}{Pro 3.1} & \multicolumn{3}{c|}{Flash 3.5} & \multicolumn{3}{c|}{Muse Spark} & DeepSeek \\
 & & low & med & high & low & med & high & low & med & high & low & med & high & low & med & high & low & med & high & low & med & high & low & med & high & low & med & high & low & med & high & \multicolumn{1}{c}{V4 Flash} \\
\bottomrule
\end{tabular}}
\end{table}

\clearpage
\section{SynthBench (reaction outcome) results, all models}

\begin{table}[h!]
\centering
\caption{Reaction-outcome accuracy (\%) by reaction class. Binary task, chance $=50$; \underline{underline} $=$ significantly above chance (one-sided binomial, $p<2.2 \cdot 10^{-4}$ after Bonferroni correction); \textbf{bold} $=$ best in column.}
\label{tab:rxn-outcome}
\tiny
\begin{tabular}{lrrrrrrrrr}
\toprule
Model & Amide & Buch. & Chan-Lam & N-Alk & Suzuki & Thio. & Urea & V.\,Leusen & \textbf{Average} \\
\midrule
\multicolumn{10}{l}{\textbf{OpenAI}} \\
GPT-5.5 (low) & 54 & 52 & 54 & 58 & 50 & 56 & 65 & 60 & 56 \\
GPT-5.5 (med) & 51 & 52 & 49 & 59 & 50 & 52 & 56 & 56 & 53 \\
GPT-5.5 (high) & 50 & 54 & 51 & 59 & 50 & 51 & 55 & 55 & 53 \\
\addlinespace[2pt]
GPT-Rosalind (low) & 50 & 49 & 51 & 55 & 50 & 51 & 50 & 61 & 52 \\
GPT-Rosalind (med) & 51 & 50 & 54 & 56 & 50 & 50 & 54 & 60 & 53 \\
GPT-Rosalind (high) & 50 & 54 & 45 & 61 & 50 & 49 & 49 & 59 & 52 \\
\bottomrule
\end{tabular}
\end{table}

\end{document}